\documentclass{ieeetj}
\usepackage{cite}
\usepackage{amsmath,amssymb,amsfonts}
\usepackage{algorithmic}
\usepackage{graphicx,color}
\usepackage{textcomp}
\usepackage{xcolor}
\usepackage{hyperref}
\usepackage{bm}

\usepackage{algorithm}
\usepackage{algorithmic}
\def\BibTeX{{\rm B\kern-.05em{\sc i\kern-.025em b}\kern-.08em
    T\kern-.1667em\lower.7ex\hbox{E}\kern-.125emX}}
\AtBeginDocument{\definecolor{tmlcncolor}{cmyk}{0.93,0.59,0.15,0.02}\definecolor{NavyBlue}{RGB}{0,86,125}}
\DeclareGraphicsExtensions{.pdf,.png,.jpg}
\usepackage{multirow}
\usepackage{caption}
\usepackage{subcaption}
\usepackage{tikz}
\usepackage[noabbrev,nameinlink,capitalize]{cleveref}
\usepackage{booktabs, multirow}
\usepackage{tabularx, array}
\let\labelindent\relax
\usepackage{enumitem}
\usepackage{comment}

\newcolumntype{L}{>{\raggedright\arraybackslash}X} 
\usepackage{fix-cm}
\usepackage{placeins}
 
\newcommand{\chg}[1]{{\color{black}#1}}
\usepackage{colortbl}

\def\OJlogo{\vspace{-4pt}$ $}
\def\seclogo{\vspace{10pt}$ $}

\def\authorrefmark#1{\ensuremath{^{\textbf{#1}}}}

\begin{document}
\receiveddate{XX}
\reviseddate{XX}
\accepteddate{XX}
\publisheddate{XX}
\currentdate{XX}
\doiinfo{XX}


\title{Sim-to-Real Transfer and Robustness Evaluation of Reinforcement Learning Control with Integrated Perception on an ASV for Floating Waste Capture}

\author{{L}uis F. W. Batista\authorrefmark{1}, Stéphanie Aravecchia\authorrefmark{1}, and Cédric Pradalier\authorrefmark{1}}
\affil{GeorgiaTech Europe - IRL2958 GT-CNRS, Metz, France}
\corresp{Corresponding author: Luis F. W. Batista}
\authornote{This research was supported by the French Agence Nationale de la Recherche, under grant ANR-23-CE23-0030 (project R3AMA).}

\begin{abstract}
Autonomous surface vessels for floating-waste removal operate under varying hydrodynamics, external disturbances, and challenging water-surface perception. We present a field-validated system that combines camera-based polarimetric perception with a lightweight DRL-based controller for floating-waste detection and capture. Camera detections are converted into water-surface target points \chg{and tracked by a} controller trained entirely in simulation and deployed directly on a retrofitted ASV platform. \chg{Our main contribution is a sim-to-real testing methodology that combines a two-stage simulation protocol with a perception abstraction module designed to mimic real camera behavior, enabling reproducible field trials and explicit evaluation of the sim-to-real gap. We apply this framework in matched simulation and field experiments across 14 disturbance regimes to expose failure modes and evaluate robustness. The results show centimeter-level terminal accuracy and indicate robust control performance under the evaluated perturbation regimes. The main source of degradation is insufficient actuation-model fidelity. We also demonstrate the system in a search-and-capture application using real camera detections in real-world conditions over areas of up to $450~m^2$. The study distills practical lessons for reliable transfer, including improved actuation-model fidelity, targeted domain randomization, and careful management of latency and timestamps across modules, while highlighting remaining challenges.}

\end{abstract}


\IEEEspecialpapernotice{\color{white}(Invited Paper)}

\maketitle

\section{INTRODUCTION}

Plastic debris in inland and coastal waters constitutes a critical sustainability challenge, with documented impacts on marine organisms, ecosystems, and human well-being~\cite{van2020plastic}. Effective mitigation requires complementary strategies that both prevent new inputs and remove legacy waste already present in the environment. Autonomous Surface Vehicles (ASVs) have emerged as capable platforms for water quality monitoring and debris retrieval~\cite{FornaiwaterSampling, chang2021autonomous}. Accordingly, autonomous cleaning tasks using ASVs are receiving sustained attention in academic~\cite{zhou2021time} and industrial~\cite{cheng2021flow} settings. Some commercial platforms are also available, such as the IADYS Jellyfishbot\footnote{\url{https://www.iadys.com/}}, which can be easily teleoperated. Increasing onboard autonomy could further support scalable operations.

\begin{figure}[tb]
    \centering
    \begin{subfigure}{0.49\linewidth}
        \includegraphics[width=\linewidth]{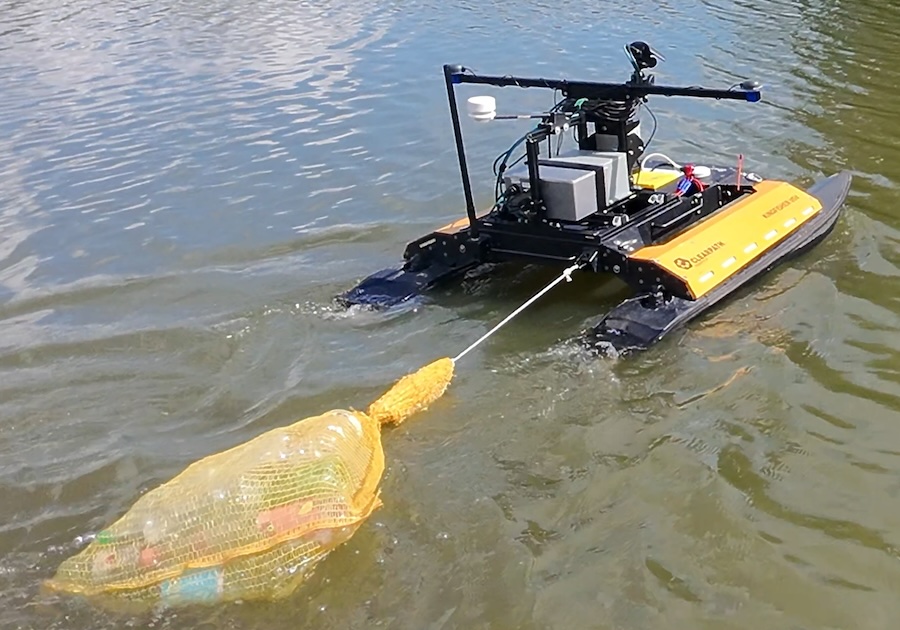}
        \caption{\textit{Kingfisher} Research Platform \chg{used in this research}}
        \label{fig:kingfisher-net}
    \end{subfigure}
    \begin{subfigure}{0.49\linewidth}
        \includegraphics[width=\linewidth]{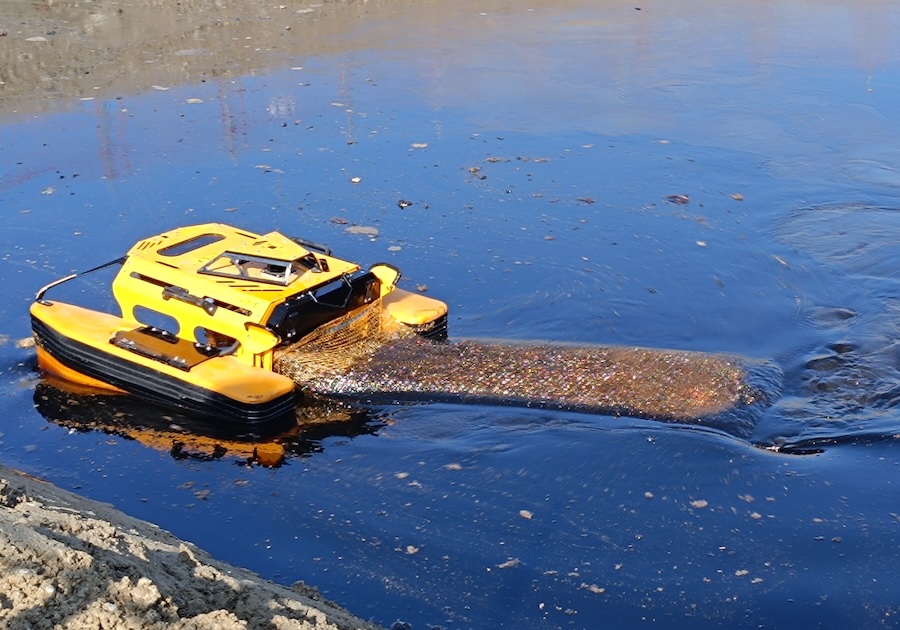}
        \caption{Jellyfishbot (Courtesy of IADYS) \chg{showing real-world payload variation}}
        \label{fig:jellyfishbot}
    \end{subfigure}
    \caption{Autonomous surface vessels fitted with nets for retrieving floating waste. The growing load from collected debris changes vessel hydrodynamics and handling.}
    \label{fig:capturing_waste}
\end{figure}

Achieving robust autonomy for floating waste removal presents several challenges. \chg{One such challenge is} payload variability during operation. As illustrated by the commercial platform shown in \cref{fig:jellyfishbot}, one practical option to capture debris is to equip the ASV with a towed net. As the net fills, changes in payload modify the vehicle’s mass, moment of inertia, and hydrodynamic drag, thereby shifting its maneuvering dynamics. Concurrently, exogenous disturbances from wind and currents act on the hull. These factors necessitate real-time control adaptation to maintain accurate and stable operation throughout the cleanup process.

Reinforcement learning (RL) techniques enable the development of control policies that adapt to model uncertainty and environmental disturbances, with growing adoption in marine robotics and ASVs~\cite{qiao2023survey}. Within this scope, point-goal navigation formulations for floating-debris interception have been explored for ASVs, showing promising robustness to environmental and payload variability~\cite{batista2024drl4asv}.

From the perception perspective, machine-learning–based object detection has been widely used to quantify floating plastics in natural waters~\cite{politikos2023using}. However, outdoor aquatic scenes exhibit variable illumination with direct sun, shadows, and specular reflections that can degrade camera-based detection. Polarization imaging can exploit the polarization state of reflected light to suppress glare and enhance object contours on the water surface. Recent results in polarimetric object detection~\cite{batista2024potato} demonstrate improved robustness to specular reflections in aquatic environments, strengthening detection reliability under challenging illumination.

\chg{
Motivated by these challenges and recent advances, we present a system that integrates polarimetric perception and RL-based control for the autonomous capture of floating waste. This work focuses on three aspects: evaluating control-policy robustness in perception-driven point-goal navigation, systematically characterizing the sim-to-real gap, and validating the complete system in autonomous field operation. In addition, through field trials, it provides a baseline demonstration of closed-loop search-and-capture that combines an onboard polarimetric perception with a RL-based control.}

\chg{Evaluating control performance when camera-based perception is part of the control loop is inherently challenging. Beyond the integrated system itself, this work contributes an evaluation methodology for precise characterization of the sim-to-real gap. Its core elements are a two-stage simulation setup and a perception abstraction module that approximates key characteristics of real camera behavior, enabling reproducible field trials and controlled simulation-to-field comparison. The resulting analysis provides practical guidance for the development and deployment of RL policies on physical platforms, particularly when perception directly influences control.}

\chg{In the context of ASV navigation, our main contributions can be summarized as:
\begin{itemize}
    \item A methodology based on two-stage simulation and a simulated perception module, enabling reproducible field trials and characterization of the sim-to-real gap.
    \item A systematic evaluation of RL-policy robustness and sim-to-real transferability under realistic disturbances.
    \item A field-validated baseline system that integrates perception and RL-based control for autonomous floating-waste capture.
\end{itemize}}

\chg{
\subsection{Outline of the paper}
\label{subsec:outline}
The paper is structured around the main components of the proposed robotic system. \cref{sec:related-work} reviews related work grouped by relevant topics, and \cref{sec:setup} describes the experimental setup, including the Kingfisher ASV, the two-stage simulation setup, and the evaluation methodology. The following three sections are largely self-contained, each dedicated to one system component, covering its formulation, methodology, results, and discussion. Accordingly, \cref{sec:polarimetric-perception} presents the polarimetric perception pipeline and formalizes the simulated perception abstraction created to enable reproducible experiments. \cref{sec:control} details the RL-based control formulation, provides an ablation study on controller robustness, and reports the corresponding sim-to-real gap metrics. \cref{sec:integration} then introduces the integrated perception-and-control system, formulates the search-and-capture problem that is evaluated both simulation and field trials. \cref{sec:discussion} and \cref{sec:conclusion} then return to the global scope of the paper, discussing overall limitations, lessons learned, future directions, and conclusions.
}

\section{RELATED WORK}
\label{sec:related-work}

Autonomous floating-waste collection is a multidisciplinary work that involves perception, control, and systems engineering. In this section, we survey correlated areas and discuss how they relate to our problem and methodology.

\subsection{Autonomous Surface Vessels and Challenges}

Autonomous Surface Vessels (ASVs) are increasingly deployed in complex, dynamic waterways, where practical autonomy faces several challenges.

One challenge is ensuring reliable control under a time-varying, partially observed payload. Roboat III~\cite{wang_roboat_2023}, an autonomous urban vessel for passenger transport, uses adaptive model predictive control (MPC) to compensate for load variations. This adaptation relies on estimating the payload from draft measurements and the resultant underwater volume. In the waste-collection setting, the payload increases as debris accumulates in the net, and the payload is not directly observable, making compensation nontrivial.

Another common challenge concerns perception. For example, Huang et al.~\cite{huang2024fieldstochasticplan} present a stochastic planning method for ASV navigation; their perception stack fuses camera and sonar to build an occupancy grid and leverages labels from the Segment Anything Model (SAM)~\cite{kirillov2023sam} to generate binary water masks and to detect shorelines. Although the application differs from debris capture, their analysis underscores the difficulties of image-based pipelines in unconstrained outdoor conditions, a limitation shared by camera-based perception in our domain.

Some studies abstract away perception and focus on ASV control. To address model inaccuracies in vessel dynamics,~\cite{lin2023asvinterception} propose a strategy to intercept a moving target ASV based on a backstepping adaptive controller with online parameter and disturbance estimation. In this work, we study an ASV designed for floating-waste collection with end-to-end integration of perception and control. Our setting shares common challenges across ASV applications, including complex dynamics, payload variability, environmental disturbances, and perception uncertainty.

\subsection{Environmental Application}

The development of ASVs for environmental applications, particularly for the task of collecting floating waste, has drawn growing interest in recent years. Many works in this area primarily emphasize the mechatronic design of the vehicles tailored for environmental monitoring or waste removal~\cite{sail2022, dash2021evaluation, smallcatamaran2015}. These efforts provide valuable contributions to platform development and environmental applications yet often lack comprehensive integration of perception and navigation suitable for autonomous operation in complex, real-world environments.

One example is the low-cost ASV for plastic waste collection presented by~\cite{owusu2024lowcostasvwaste}. While the system is purpose-built for waste retrieval, it focus on affordability and mechanical design, with limited evaluation of autonomy or perception-driven navigation capabilities

\chg{The cleanup application is also addressed in prior work at the planning level}. Sudha et al.\cite{sudha2025informativepp} combine perception-based target detection and tracking with informative path planning, \chg{while Barrionuevo et al.\cite{barrionuevo2025optimizing} propose a simulation-only DRL framework for cooperative plastic-waste collection with heterogeneous ASVs. While both consider related cleanup scenarios, their focus is on higher-level planning and coordination}.
In contrast, we prioritize quantitative robustness evaluation of a RL-based control pipeline, including the coupling from perception to control.

\begin{figure*}[htb]
  \begin{subfigure}[b]{0.325\linewidth}
    \centering
    \includegraphics[width=\linewidth]{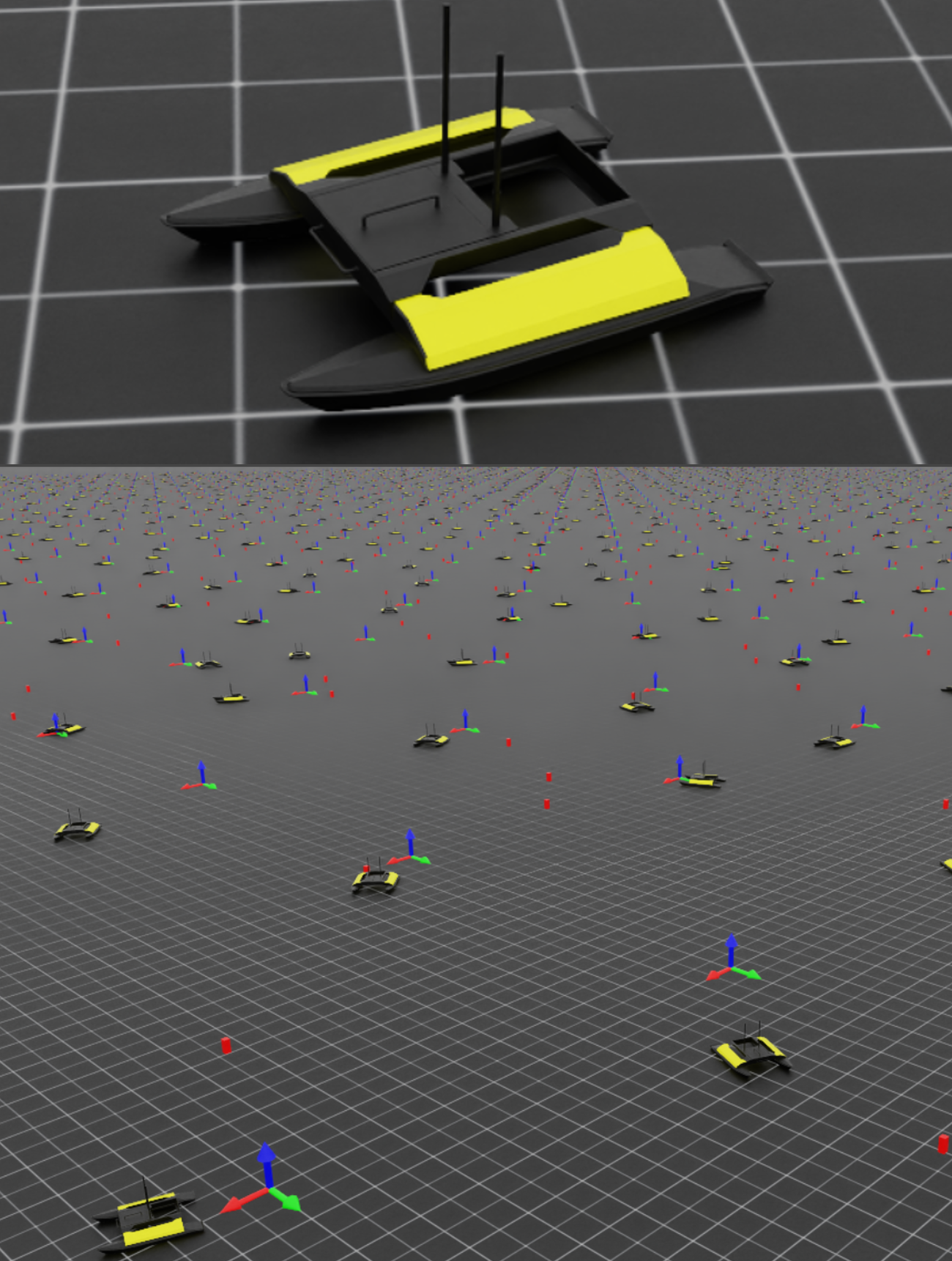}
    \caption{Isaac Lab}
    \label{fig:env-isaac}
  \end{subfigure} \hfill
  \begin{subfigure}[b]{0.325\linewidth}
    \centering
    \includegraphics[width=\linewidth]{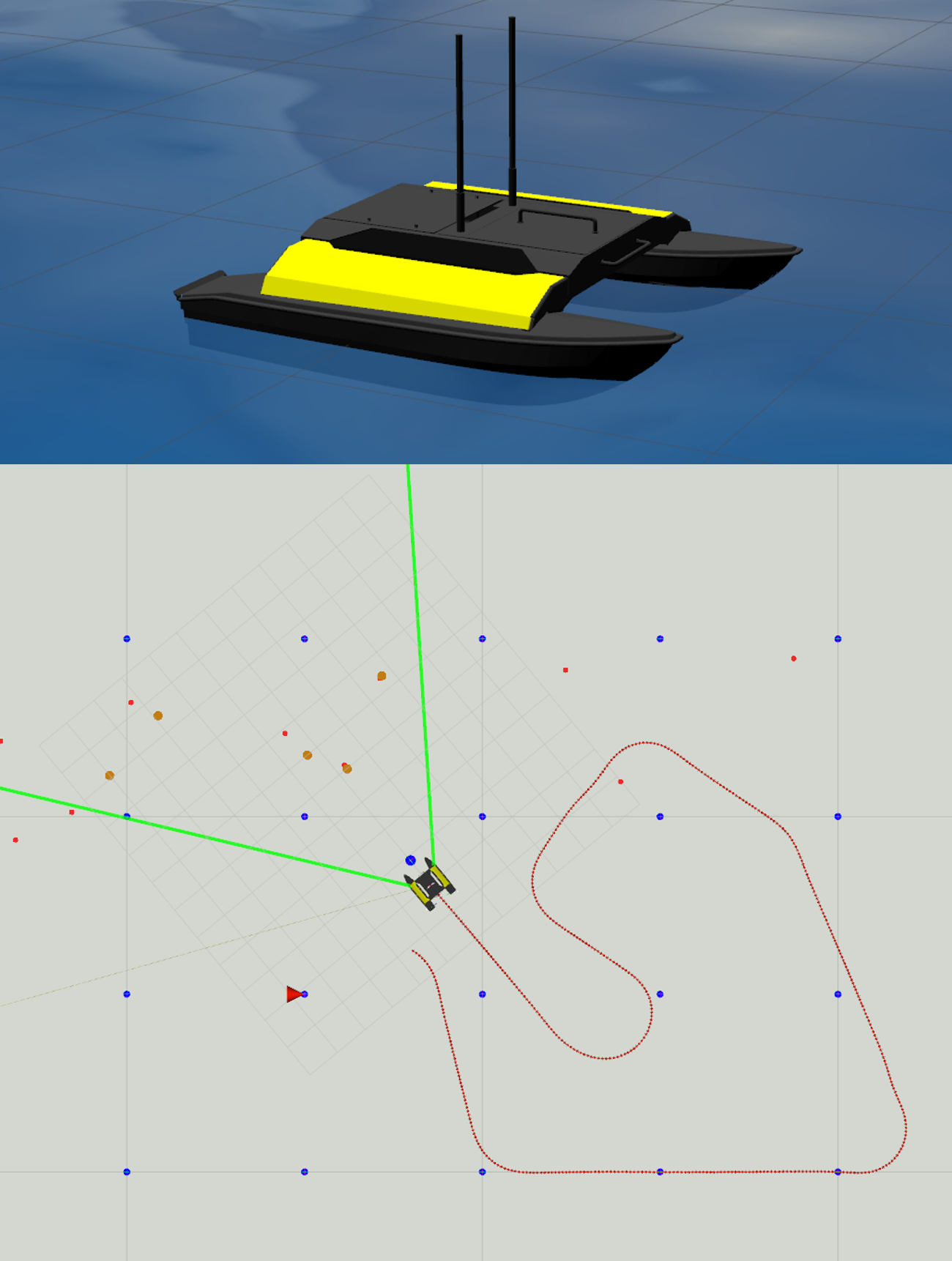}
    \caption{Gazebo Simulation}
    \label{fig:env-gazebo}
  \end{subfigure} \hfill
  \begin{subfigure}[b]{0.325\linewidth}
    \centering
    \includegraphics[width=\linewidth]{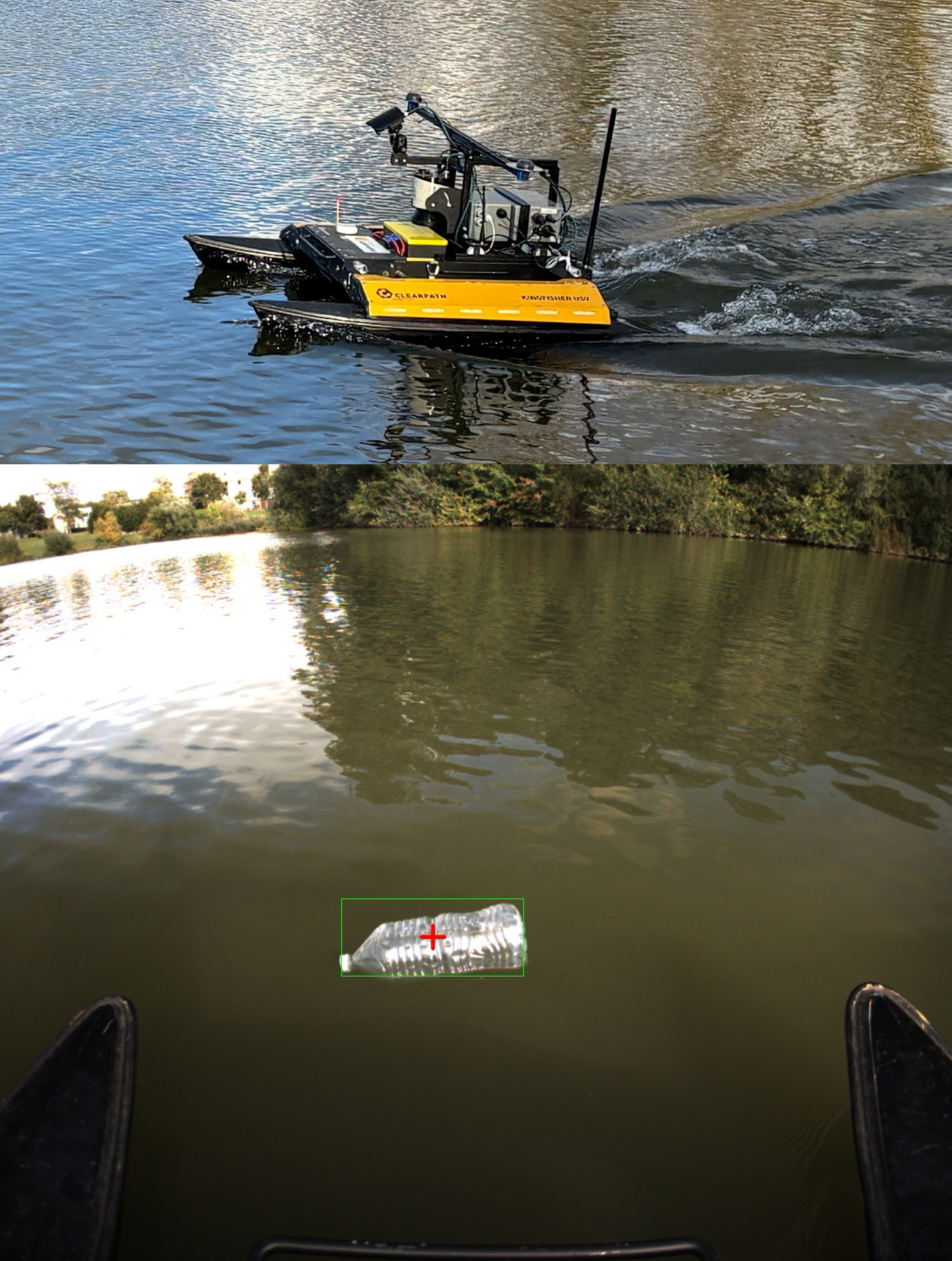}
    \caption{Robot Platform}
    \label{fig:env-real}
  \end{subfigure}
  \caption{Three environments used in this work: (a) platform for end-to-end evaluation in field trials; (b) testbed for integration and robustness evaluation in simulation; (c) highly-parallel environment used for policy training.}
  \label{fig:env-exp-setup}
\end{figure*}

\subsection{Perception \chg{for Floating Debris}}

Floating-waste identification can be posed as an object-detection problem. Deep learning has advanced perception substantially in recent years~\cite{zou2023object} and has been applied to marine litter detection. \chg{Politikos et al.~\cite{politikos2023using} compiled a large database of published studies that use AI to process macrolitter datasets and identified that} existing pipelines rely on satellite imagery, drone footage, or bridge-mounted cameras, which are not readily deployable on robotic platforms designed to collect debris. \chg{Although several litter datasets exist~\cite{cheng2021flow,jmse13081560}}, current detectors often exhibit limited cross-domain generalization on natural water surfaces~\cite{jia2023deep}, highlighting the need for models that remain robust across diverse conditions.

Detecting floating objects is complicated by outdoor illumination variability, sun glare, and specular reflections from the water surface~\cite{zhang2021survey}. To mitigate these effects, recent work has explored multimodal sensor fusion, such as pairing a long-wave infrared polarimetric camera with a visible-band optical sensor~\cite{iqbal2022object}, or combining millimeter-wave radar with a camera~\cite{cheng2021flow}. However, achieving reliable temporal synchronization and spatial correspondence across modalities introduces additional system complexity and may adversely affect overall detection performance.

Advances in microgrid polarimetric sensors have enabled the use of polarization cues for a range of vision tasks, including object detection. Blin et al. demonstrated the utility of polarimetric imagery under adverse illumination and examined several multimodal fusion strategies~\cite{blin2019adapted, blin2021multimodal}. While the reported results are promising, the implementation relies on two independent sensors without pixel-wise correspondence and require careful temporal and spatial calibration, which increases system complexity.

For water-surface applications, a dedicated dataset with polarimetric imagery was introduced to address the challenge of detecting floating debris in outdoor conditions~\cite{batista2024potato}. Empirical results showed that leveraging polarization to suppress water-surface reflections improves detection accuracy. The same dataset was later used for semantic segmentation, with studies comparing multiple fusion strategies and reporting increased performance (and inference time) as additional modalities are incorporated~\cite{batista2025evalpolfusion}. In our setting, a model that outputs the approximate center position of a target is sufficient. With that in mind, our study \chg{employs an} object detection pipeline that leverages polarimetric cameras to obtain reflection-attenuated images.

\subsection{Reinforcement Learning-based Control \chg{for ASVs}}
\label{subsec:relatedwork-rl-control}

Deep reinforcement learning (DRL) has demonstrated robust real-world control across mobile platforms including quadrupeds, full-size humanoids, and vision-based quadrotors~\cite{hoeller2024anymal, radosavovic2024humanoid, kaufmann2023championdrone}. In contrast, DRL for ASVs remains in early stages with limited hardware validation. According to Qiao et al.~\cite{qiao2023survey}, adoption is constrained by the difficulty of accurately modeling nonlinear hydrodynamic interactions and environmental disturbances.

DRL-based control for ASVs has been studied predominantly in simulation. Zhao et al.~\cite{zhao2020path} demonstrate a DRL controller for path following that remains effective under complex system dynamics. Similarly, Lin et al.~\cite{lin2023robust} propose an IQN-based local planner that compensates for unknown currents while performing obstacle avoidance. Additional RL formulations for ASV control have been reported~\cite{zhang2021model, wang2020reinforcement, wang2022reinforcement}. However, these studies largely remain confined to simulation, leaving sim-to-real transfer insufficiently validated.

\chg{Prior studies have benchmarked DRL-based marine control against traditional baselines and reported improvements in tracking and disturbance rejection, including comparisons with PID~\cite{wang2023path}, MPC~\cite{batista2025drlrobustness}, and NMPC-based~\cite{wang2023deep} controllers. Taken together, these findings indicate that DRL is a promising control paradigm for marine systems, with potential benefits in tracking accuracy, disturbance rejection, and, in some cases, auxiliary objectives such as energy efficiency. However, our work does not revisit comparisons with classical controllers, and instead focuses on the transferability and robustness of RL policies.}

\chg{Although field evaluations of DRL controllers for ASVs are becoming more common, real-world validation remains limited across task domains. Existing studies are largely early-stage or centered on path-tracking~\cite{WOO2019155, DERAJ2023113937, SLAWIK202421}, with other ASV tasks receiving comparatively less attention.}

\chg{Among these underexplored tasks, floating-waste capture can be formulated} as point-goal navigation task without a terminal-velocity constraint. Similar tasks have been investigated in other domains~\cite{el2023drift}. Specifically for ASVs,~\cite{batista2024drl4asv} introduced a software framework for developing capture tasks, and reported preliminary field experiments. A subsequent study evaluated the robustness of DRL policies in field trials~\cite{batista2025drlrobustness}, but perception was not considered in those experiments. In our work, we perform real-world testing that includes camera-based perception in the loop with a focus on evaluating the robustness of the trained policy.

\subsection{Robustness and Policy Transfer}\label{sec:sota-robustness}

We aim to train a policy that remains robust to external disturbances. In the literature, the robustness of RL agents is often intrinsically linked to uncertainty. As shown in~\cite{lockwood_review_2022,zhang_robust_2020}, each element of the Markov Decision Process (MDP) is subject to uncertainty: states, actions, transition dynamics, and rewards. Following the taxonomy in~\cite{da_survey_2025}, we employ domain randomization (DR) over observations, actions, and dynamics, and apply reward shaping to stabilize learning and discourage brittle behaviors (see \cref{sec:control-methodo}).

However, even robust policies may fail to transfer to real hardware and the resulting performance drop is known as the sim-to-real gap. \chg{The sim-to-real gap remains a central challenge in robotics and continues to limit reliable policy transfer. Among the strategies proposed to mitigate it, domain randomization (DR) has emerged as one of the most widely adopted approaches for improving robustness and transferability~\cite{hofer2021simtoreal-techniques}.} Complementary surveys~\cite{salvato2021crossings2rg,hu2023digitaltwin} attribute this gap to mismatches between simulation and the robot in sensing, actuation, and unmodeled dynamics. In this paper, we evaluate both aspects jointly: we quantify the sim-to-real gap on hardware and probe robustness through controlled perturbations of each element of the MDP.

\chg{In the maritime vehicle domain, Zheng et al.~\cite{zheng2025simtoreal} propose a sim-to-real transfer framework that studies multi-factor domain randomization and policy transferability. They introduce a separate \emph{pseudo-real} environment as an independent sim-to-sim test stage to emulate and investigate the reality gap under controlled conditions. Although conceptually related to our two-stage simulation methodology, our objective is different: we use the two-stage pipeline for system-level integration and preliminary transfer assessment, while the second simulation stage enables matched simulation--field experiments to evaluate the sim-to-real gap for real-world deployment.}

\chg{A further challenge in our setting is that transfer must be analyzed with perception in the loop. Many sim-to-real studies emphasize dynamics and actuation mismatches, but in vision-based control the quality and structure of the perceptual interface can also strongly affect transfer.}
To address this, we draw inspiration from digital-twin methodologies in autonomous driving~\cite{hu2023digitaltwin} and introduce the simulated perception abstraction described in \cref{sec:polarimetric-perception}. This lightweight interface operates at the perception--control boundary \chg{and can be used consistently in both simulation and field experiments, enabling reproducible component-level validation and more precise analysis of failures attributable to dynamics rather than perception.}

Aiming toward zero-shot transfer of a robust policy, our work \chg{extends these ideas into a structured evaluation methodology. We assess transferability through matched simulation--field experiments, enabling the identification of failure sources and the targeted improvement of simulation fidelity.}

\section{EXPERIMENTAL SETUP}
\label{sec:setup}

In this section we describe the three environments that are used in this work, illustrated in \cref{fig:env-exp-setup}. Specifically, we implement an environment based on Isaac Lab to train the reinforcement learning policies; a Gazebo-based simulation for integration and validation; and a \textit{Kingfisher} ASV research platform for field trials. Each environment is detailed in a dedicated subsection.

\begin{figure*}[t!]
    \centering
    \begin{subfigure}[b]{0.99\textwidth}
        \centering
        \includegraphics[width=\linewidth]{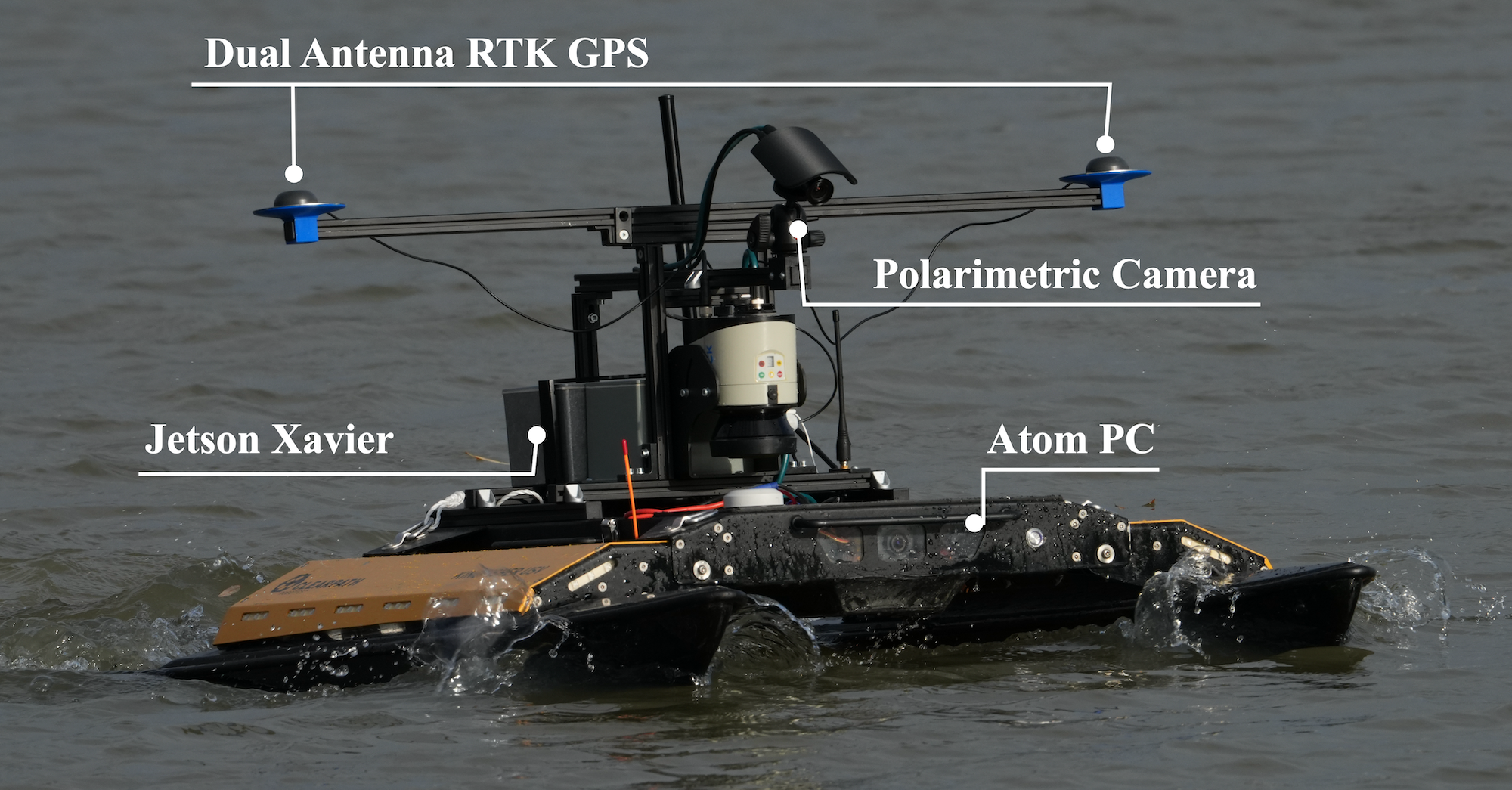}
        \caption{ASV Overview}
        \label{fig:kingfisher-overview}
    \end{subfigure}
    
    \vspace{1em} 

    \begin{subfigure}[b]{0.535\textwidth}
        \centering
        \includegraphics[width=\linewidth]{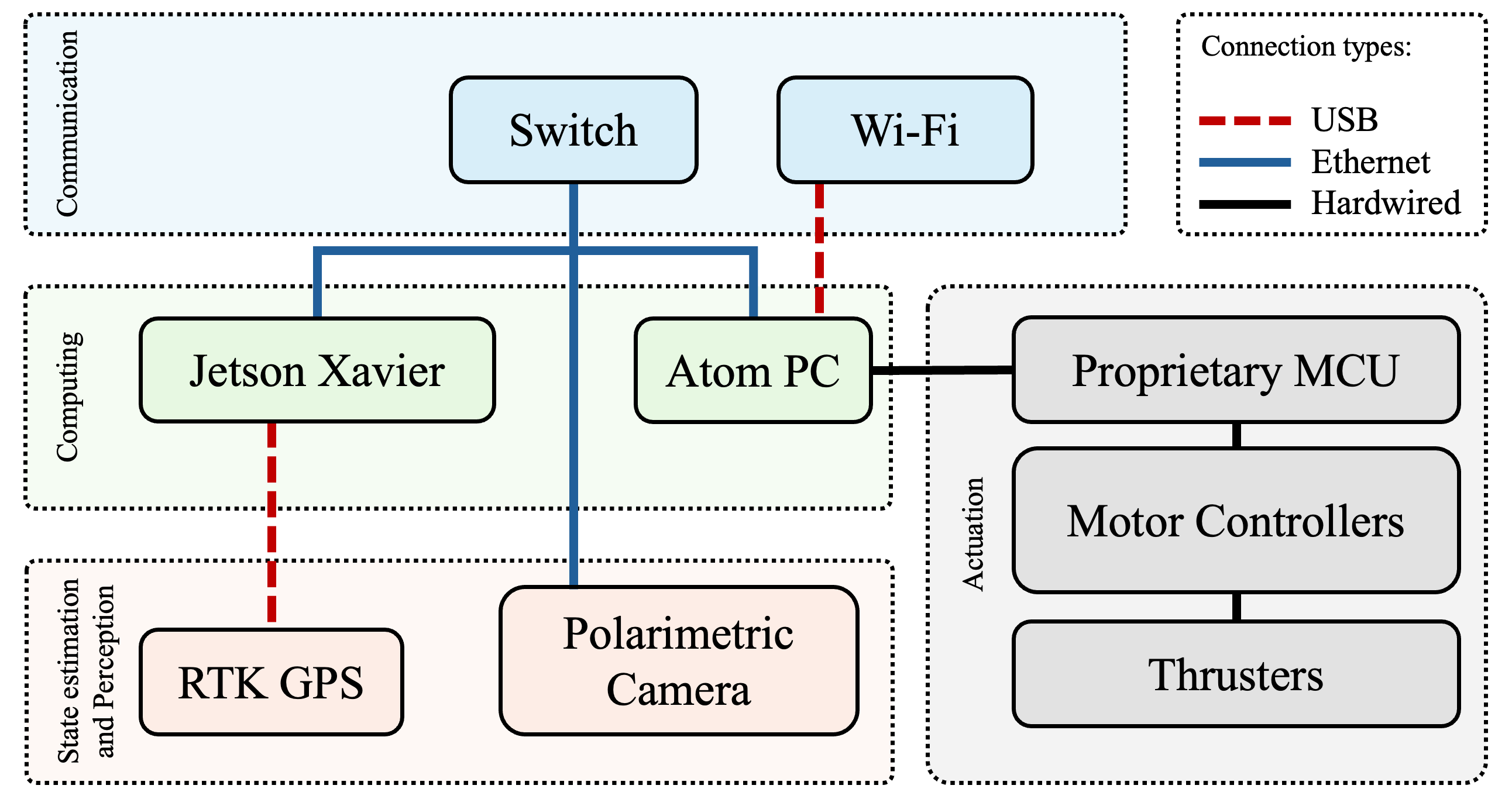}
        \caption{Block diagram highlighting main components \chg{and connection types}}
        \label{fig:electrical}
    \end{subfigure}
    \hfill
    \begin{subfigure}[b]{0.455\textwidth}
        \centering
        \includegraphics[width=\linewidth]{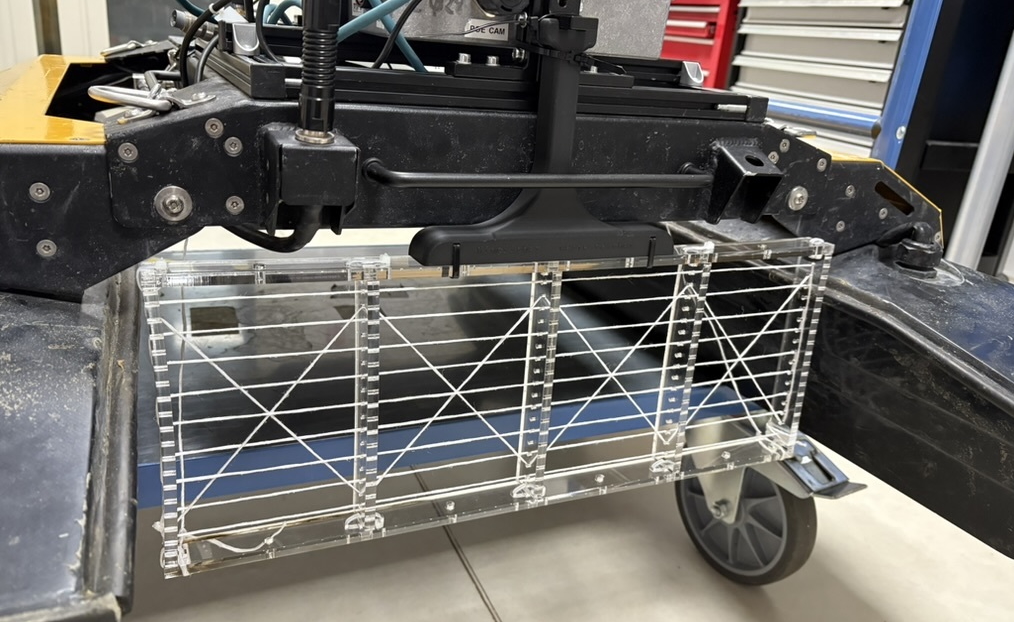}
        \caption{Frame attached between the hulls}
        \label{fig:bottle-catcher}
    \end{subfigure}

    \caption{The \textit{Clearpath Kingfisher} ASV used in field experiments. 
    The ASV is equipped with a diverse set of sensors, including RTK-GPS and a polarimetric camera. 
    It contains two embedded computers for data processing: an Atom PC and a Jetson Xavier GPU. 
    The main architecture and system connections are presented in (b). 
    When collecting floating waste, a custom-design frame was attached in the bottom of the ASV as depicted in (c).}
    \label{fig:hardware-platform}
\end{figure*}

\subsection{Robot Platform}
\label{subsec:hardware-platform}

We conducted field experiments with the \textit{Kingfisher} ASV (Clearpath Robotics), a compact catamaran and predecessor to the \textit{Heron}. The hull measures $1.35\,\text{m}$ (length) by $0.98\,\text{m}$ (beam) with a total mass of $35.82\,\text{kg}$. Propulsion is provided by two independently driven thrusters (one per hull), each powered by a $400\,\text{W}$ motor. The platform was substantially retrofitted relative to the stock configuration, including replacement of propeller blades, an additional GPU-enabled embedded computer, and additional sensors, as shown in \cref{fig:hardware-platform}.

The system is powered by two $22\,\text{Ah}$, 4-cell LiPo batteries, providing over 5 hours of operation in our field trials. A DC--DC step-down converter supplies a regulated $12\,\text{V}$ rail. Although the ASV carries a monocular camera and a 2D lidar, these were not used in the experiments reported here; instead, we use a Triton $5.0\,\text{MP}$ (TRI050S1\text{-}QC) polarimetric camera built with a Sony IMX264MYR sensor, recording color and polarization at $2448\times 2048$ pixels. The unit is paired with a $6\,\text{mm}$ lens yielding an angle of view of $80.8^{\circ}(H)\times 61.6^{\circ}(V)$ and is mounted on the \textit{Kingfisher} facing forward with a \chg{downward pitch of approximately 37 degrees} to maximize visibility of floating objects up to close approach (see \cref{fig:hardware-platform}). State estimation relies on an SBG Ellipse-D IMU equipped with a dual-antenna RTK-GNSS receiver, achieving position, velocity, and heading accuracies of $0.02\,\text{m}$, $0.03\,\text{m/s}$, and $0.5^{\circ}$, respectively.

Thrust input commands are in the $[-1,\,1]$ range. Input commands and thruster force mapping was identified via bollard-pull testing, \chg{a static test in which the thrust generated at fixed commands is measured,} resulting in the curve presented in \cref{fig:thruster-force-curve}. The input commands must be provided at a rate of at least $10\,\text{Hz}$ and are then processed by a proprietary microcontroller unit (MCU), which outputs PPM (Pulse-Position Modulation) signals to the motor controllers. Probing the PPM signals revealed that the MCU implements a software rate limiter that constrains command slew and introduces a temporal response shown in \cref{fig:cmd-drive-response}.

Tasks are distributed between two onboard computers: the Atom PC and the Jetson Xavier AGX\chg{, both using Ubuntu 20.04 LTS}. Time synchronization between the two uses NTP during startup. We use the robot operating system (ROS) \chg{Noetic} framework~\cite{quigley2009ros} to integrate our components. The Atom PC is connected to the MCU and hosts the ROS master. It handles low-level control, such as issuing thruster commands \chg{using a modified version of Clearpath’s Kingfisher software stack that was adapted to our platform modifications}.
The Jetson Xavier handles resource-intensive tasks, including raw image processing \chg{implemented as a custom Python ROS node that leverages NumPy vectorization}, object-detection inference \chg{is executed on the GPU using the trained model exported to ONNX and optimized with TensorRT}, RL policy inference \chg{is executed in a custom Python node on the CPU due to the small MLP architecture}. 
Access to the ASV during field tests is provided via Wi-Fi, enabling remote experiment start, stop, and monitoring.

\subsection{Gazebo Simulation}\label{subsec:simulation-environment}

We employ a simulation stack based on Gazebo with the UUV Simulator plugin~\cite{bingham2019toward} to complement the Isaac Lab-based training environment presented in \cref{subsec:training-environment}.
This setup provides an independent physics and hydrodynamics implementation, enabling an initial validation of policy behavior and checking against potential overfitting to the custom dynamics implementation in Isaac Sim. It allows to quickly assess consistency of learned behavior before field tests. 

Additionally, the Gazebo simulation stack was customized to include controllable perception-in-the-loop configuration. 
Instead of a physical camera, we instantiate a calibrated virtual sensor using the intrinsics from camera calibration parameters and a deterministic projection model (\cref{subsec:pp-calib-proj}). Within this pipeline, we can inject controllable detection noise and projection errors to probe robustness to perception bias and uncertainty during policy evaluation in simulation.

\begin{figure}[htb]
    \centering
    \begin{subfigure}[b]{0.99\linewidth}
        \centering
        \includegraphics[width=\linewidth]{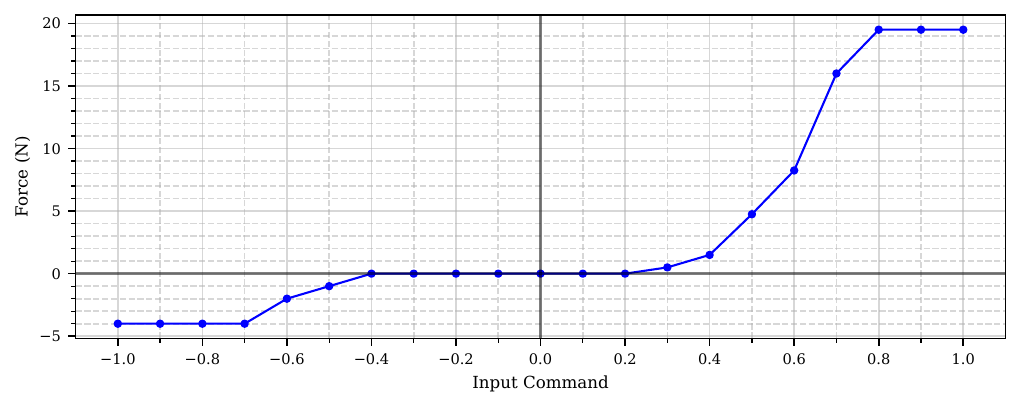}
        \caption{Measured thruster force curve obtained via a bollard-pull test.}
        \label{fig:thruster-force-curve}
    \end{subfigure}
    \begin{subfigure}[b]{0.99\linewidth}
        \centering
        \includegraphics[width=\linewidth]{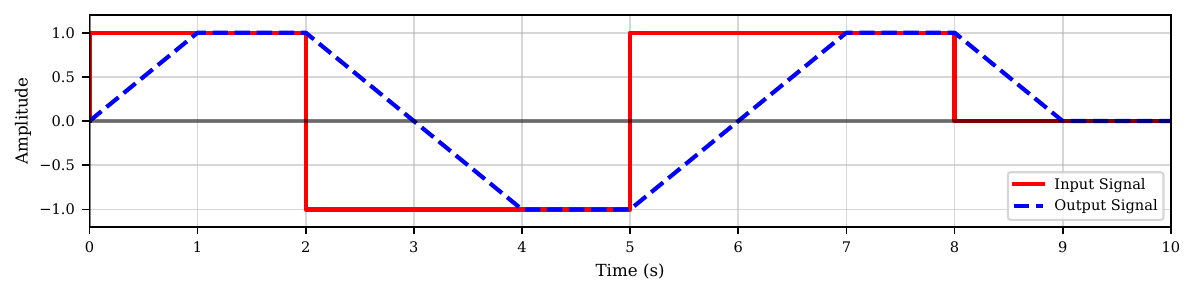}
        \caption{Measured MCU command limiting. A full-scale transition from \(-1\) to \(+1\) requires up to \(\sim 2\,\mathrm{s}\), introducing a significant lag between input and motor controller command.}
        \label{fig:cmd-drive-response}
    \end{subfigure}
    \caption{Thruster actuation characterization: (a) static force curve; (b) dynamic command response.}
    \label{fig:thruster-characterization}
\end{figure}

The UUV Simulator plugin is updated using the hydrodynamic coefficients in the linear/quadratic damping model (\cref{eq:damping}) and the thruster force curve obtained via system identification experiments on the \textit{Kingfisher} ASV (\cref{fig:thruster-force-curve}). The software rate-limit identified in the MCU is replicated in software and introduces the same response shown in \cref{fig:cmd-drive-response}.

The Gazebo environment used for these experiments is illustrated in \cref{fig:env-gazebo}, including the model, simulated targets, and the simulated perception pipeline.

\subsection{Isaac Lab}\label{subsec:training-environment}

For training the RL policies, we use an environment based on Isaac Lab~\cite{mittal2023orbit} due to its batched simulation paradigm that allows massive parallelization.

ASV simulator has already been integrated into \textit{OmniIsaacGym}~\cite{batista2024drl4asv,makoviychuk2021isaac}, allowing for GPU acceleration across many independent rollouts, enabling thousands of environments to be advanced simultaneously. This degree of parallelism substantially increases sample throughput and reduces wall-clock training time, making model-free policy optimization practical for our setting. In our work, we implement the core dynamics necessary for ASV simulation (hydrostatics, hydrodynamics, and thruster actuation) on the newer \textit{Isaac Lab 2.1}. \Cref{fig:env-exp-setup} (c) illustrates the environment used for policy training. It is implemented in a way that it accepts the same hydrostatic and hydrodynamic coefficients as the Gazebo-based environment, ensuring parameter parity across platforms. This includes an identical thruster model with an optional software rate limiter to reproduce command-rate constraints. During training, these coefficients and actuation behaviors are exposed to domain randomization, enabling variability while preserving physical consistency. Details of policy training are presented in \cref{sec:control}.

\chg{
\subsection{Evaluation Methodology}
\label{subsec:simulated-perception-intro}

A central objective of this work is to ensure robust sim-to-real transfer when perception is part of the control loop. Direct comparison with real perception is challenging because simulation cannot faithfully capture the complexity of outdoor illumination and scene-dependent variability. This, in turn, complicates a controlled assessment of control policy transferability. To address this limitation, we introduce a \emph{simulated perception} (SP) interface, illustrated in \cref{fig:diagram-perception}.

The simulated perception is defined at the perception--control boundary. Rather than estimating detections from images, it generates the goal input to the control policy from a programmable target position using the calibrated camera model with controllable perturbations. This abstraction preserves the controller interface, avoids the need for photorealistic image simulation, and enables repeatable, fine-grained evaluation of perception errors across simulation and field experiments. \Cref{subsec:perception-pipeline} presents a formal definition and implementation details.

\begin{figure}[htb]
    \centering
    \includegraphics[width=\linewidth]{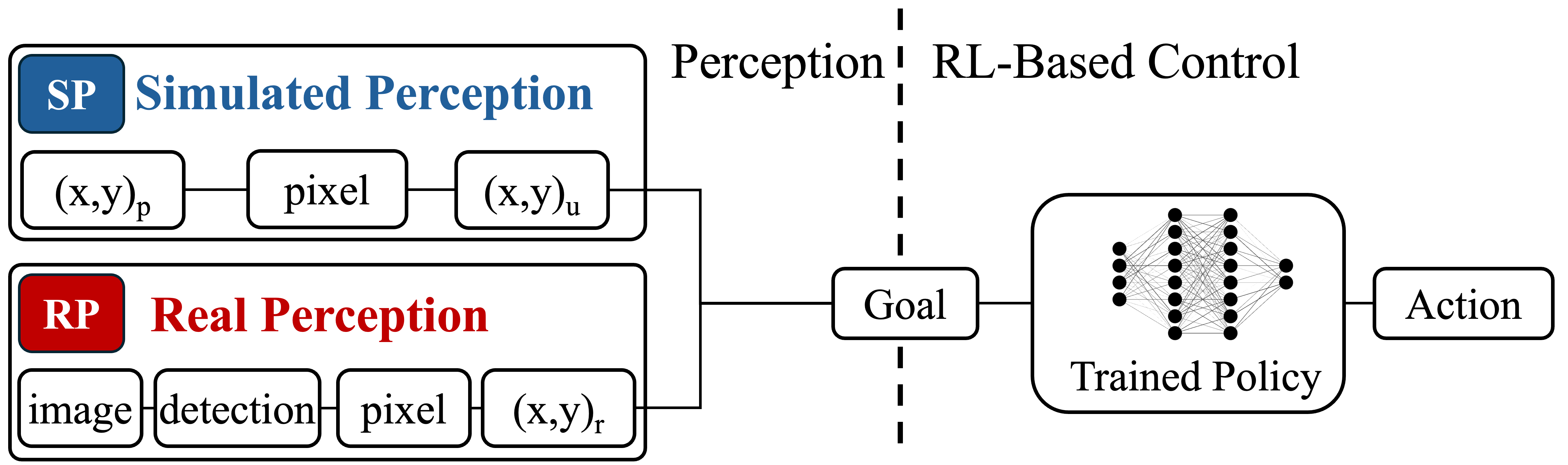}
    \caption{\chg{Perception and RL-based control pipeline. Both simulated perception (SP) and real perception (RP) can  be used to provide goal inputs to the controller. SP generates this input from a programmable target position with controllable perturbations, whereas RP obtains it from the onboard polarimetric camera and detector.}}
    \label{fig:diagram-perception}
\end{figure}

This simulated perception is central to the evaluation protocol represented in \cref{fig:diagram-test}, which follows a progressive validation strategy. (1)~The RL policy is first trained in Isaac Lab. (2)~Simulated perception is then used to reproduce the same goal configurations in Gazebo and on the real ASV, isolating transfer effects mainly due to dynamics, actuation, and timing while removing image-based detection variability. (3)~Finally, the integrated system is evaluated in field trials with real perception (RP). The difference between matched SP experiments in simulation and field is used to quantify the sim-to-real gap, whereas RP experiments assess end-to-end autonomy under realistic conditions.

\begin{figure}[htb]
    \centering
    \includegraphics[width=\linewidth]{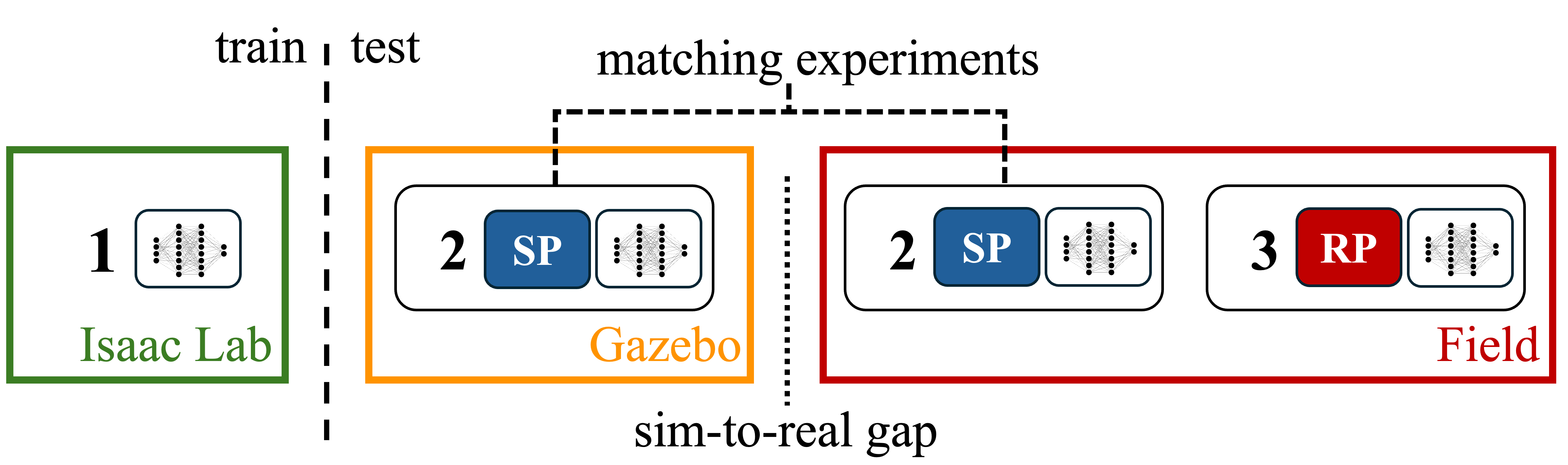}
    \caption{\chg{Progressive training and evaluation protocol with increasing system realism. (1) The policy is trained in Isaac Lab. (2) Matched experiments with simulated perception are performed in Gazebo and on the real ASV to quantify the sim-to-real gap while keeping the perception interface fixed. (3) Field trials with real perception evaluate the fully integrated system.}}
    \label{fig:diagram-test}
\end{figure}

Accordingly, the simulated perception is used as the main evaluation instrument in the control experiments of \cref{sec:control} and as an intermediate validation stage before the fully integrated perception-driven trials presented in \cref{sec:integration}. This methodology enables controlled and reproducible analysis of failure modes while preserving a direct path toward real-world deployment.
}

\section{POLARIMETRIC PERCEPTION}
\label{sec:polarimetric-perception}

This section focuses on the perception pipeline. The goal is to detect floating objects in the polarimetric images, and map the detections to a goal in the robot frame. The detections are produced in the image plane and its pixel coordinates are projected onto the water surface, yielding a planar goal in the robot (ASV) frame $\mathbf{g}_r = [x_r\, y_r]^\top$ for the downstream controller (see \cref{sec:control}). This goal-based abstraction reduces the dimensionality of the observation and simplifies the coupling between perception and control modules.
It decouples camera choice and facilitates testing components independently.

\subsection{Polarimetric Object Detection}\label{subsec:polarimetric-object-detection}
We use the polarimetric camera described in \cref{subsec:hardware-platform} to obtain color images with reduced surface specular reflections. This design choice is motivated by the results presented in~\cite{batista2024potato}, which reports that polarimetric sensing improves detection precision in water-rich outdoor environments.

\begin{figure}[htb]
    \centering
    \begin{subfigure}[b]{0.493\linewidth}
        \centering
        \includegraphics[width=\linewidth]{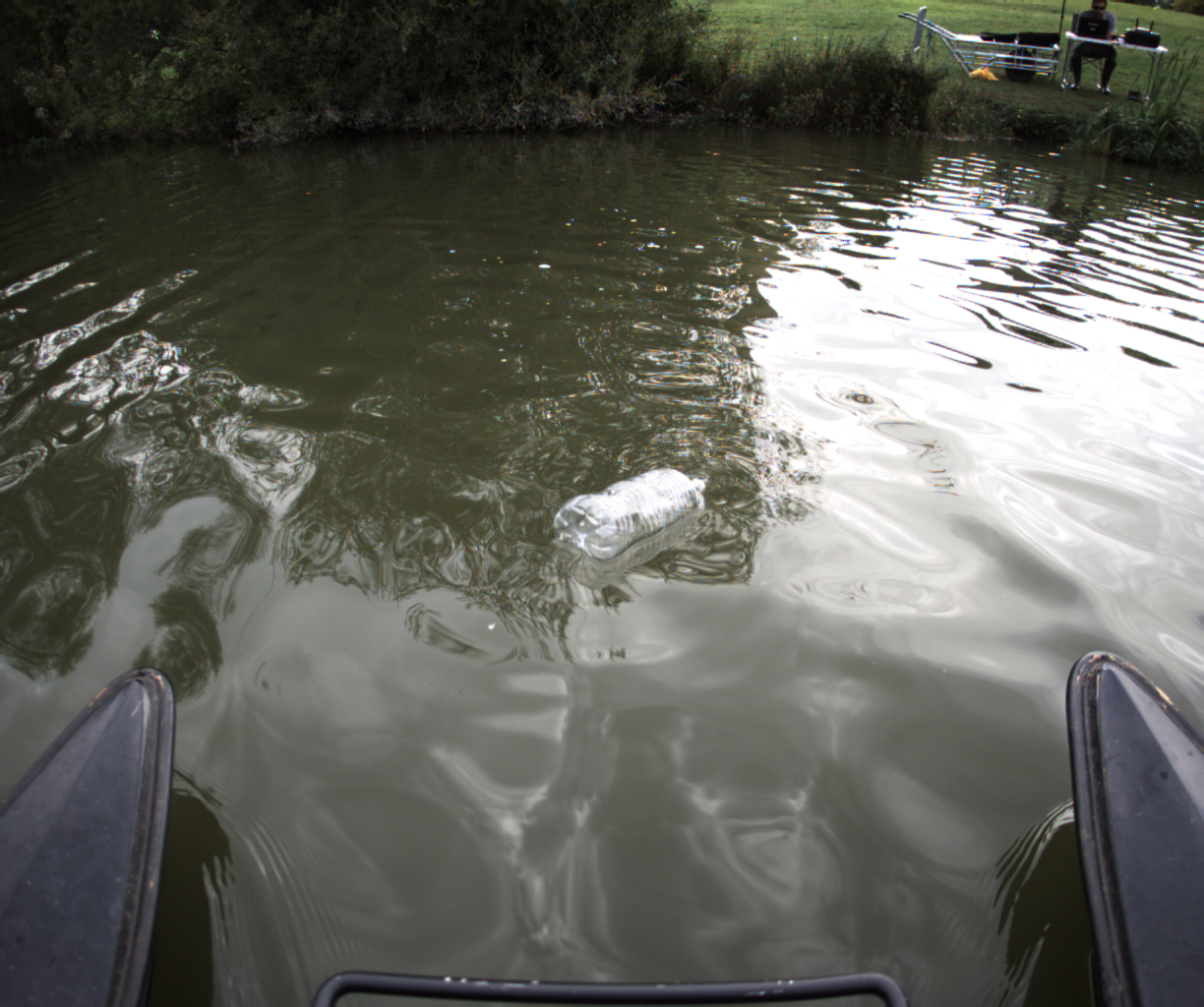}
    \end{subfigure}
    \begin{subfigure}[b]{0.493\linewidth}
        \centering\includegraphics[width=\linewidth]{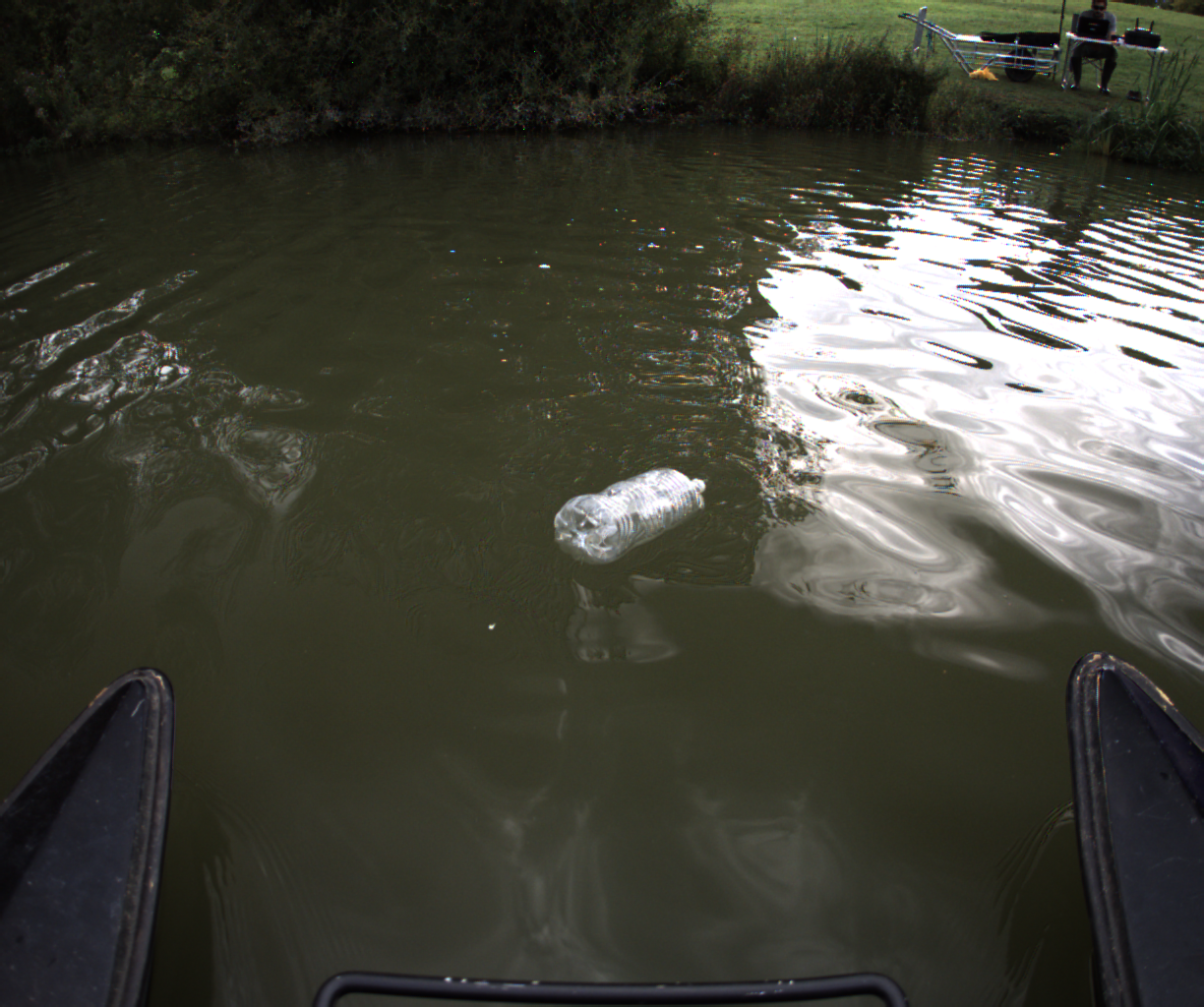}
    \end{subfigure}
    \begin{subfigure}[b]{0.493\linewidth}
        \centering
        \includegraphics[width=\linewidth]{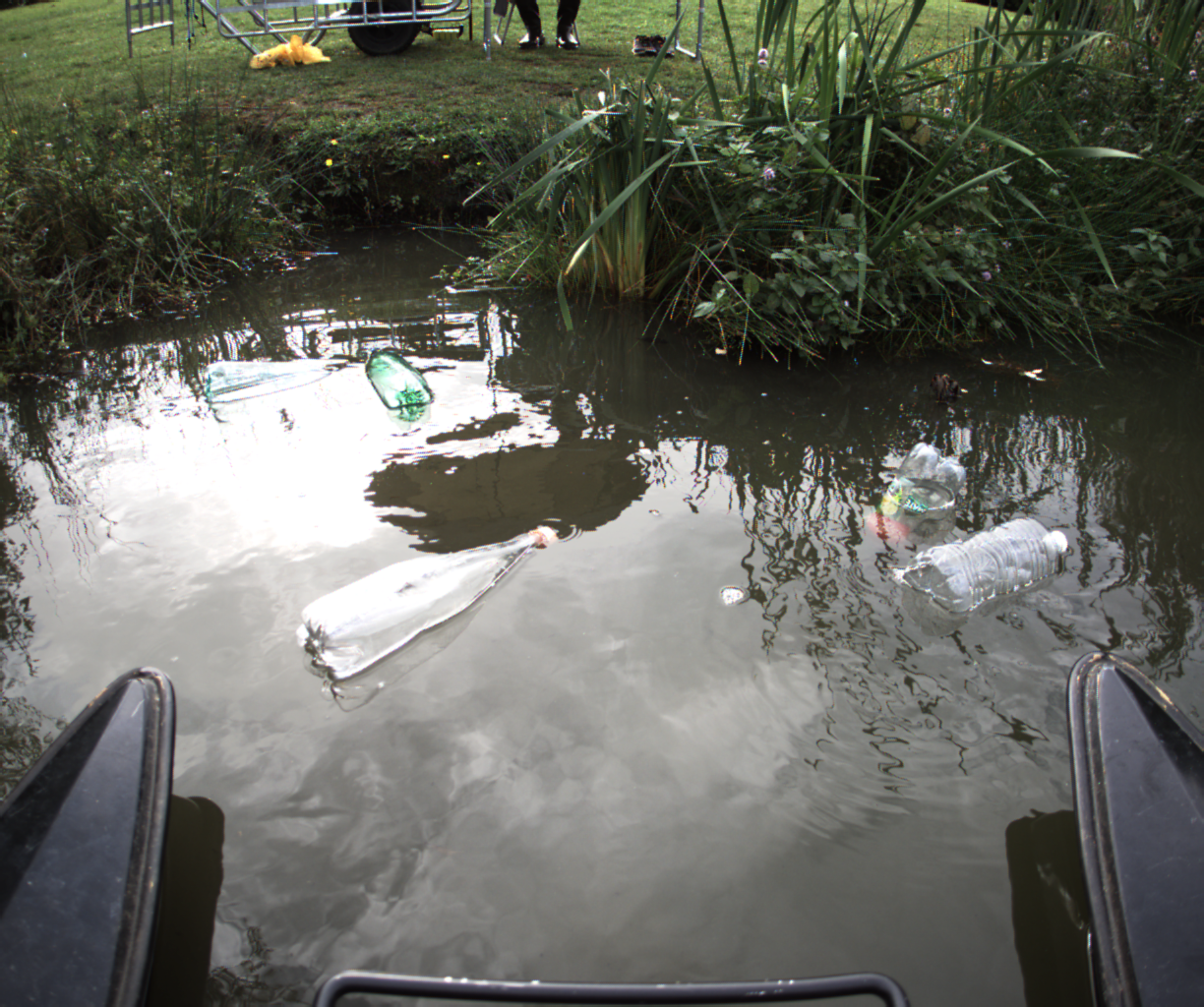}
        \caption{Normal RGB images}
    \end{subfigure}
    \begin{subfigure}[b]{0.493\linewidth}
        \centering
        \includegraphics[width=\linewidth]{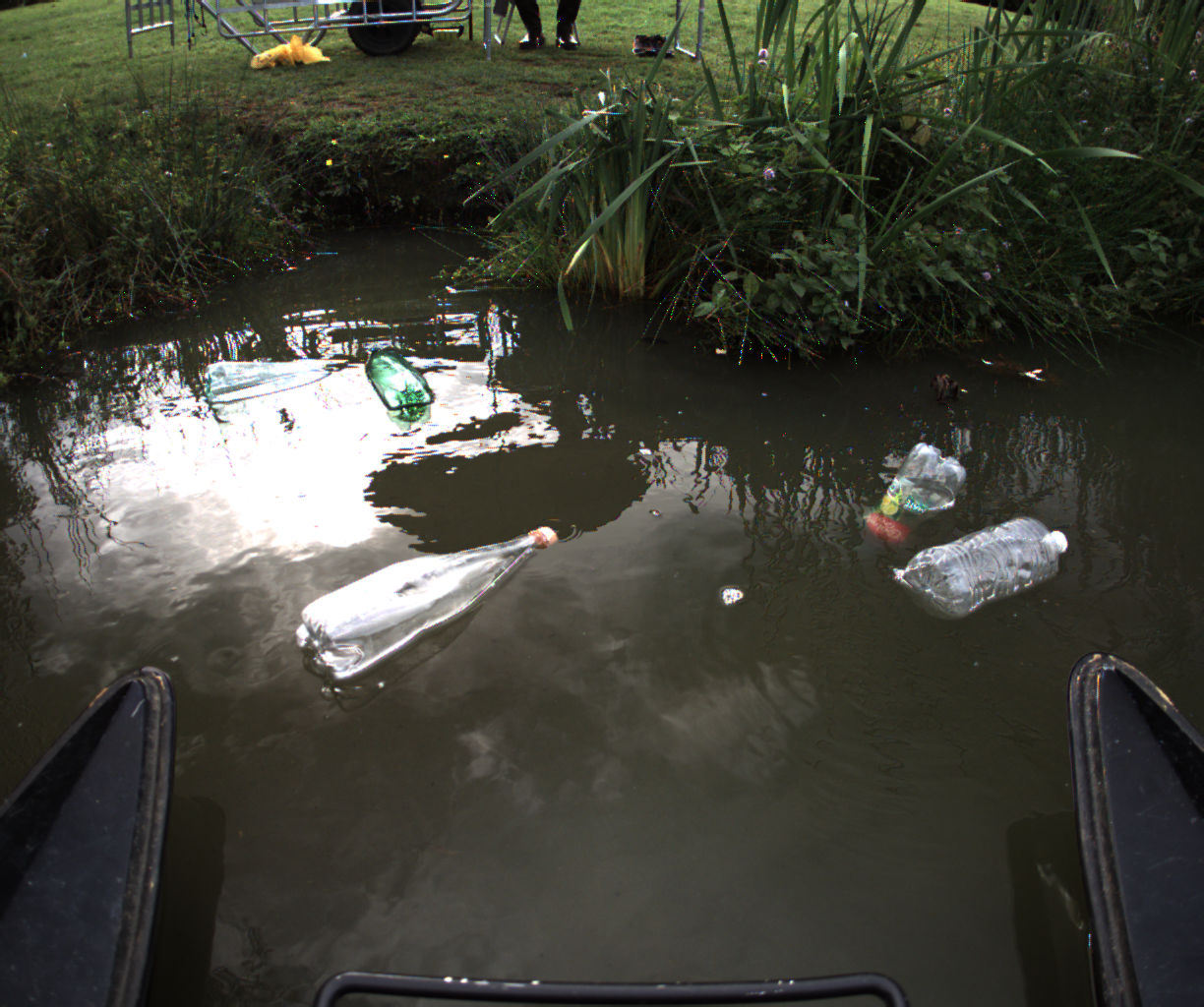}
        \caption{Polarimetric-enhanced images}
    \end{subfigure}
    \caption{Example of images generated by the polarimetric camera. The polarization information can be used to reduce the reflections on the water surface.}
    \label{fig:polarimetric-image-samples}
\end{figure}

To use the polarimetric camera, we first \chg{process the raw image to recover either the color or the polarimetric image.}
Our implementation follows the extraction pipeline detailed in~\cite{batista2024potato} and \cref{fig:polarimetric-image-samples} shows outputs of RGB and polarimetric-enhanced images. In this work, we perform detection only on the glare-suppressed images which are used through all of our field experiments.

We detect floating waste using \emph{YOLOv5-m}~\cite{yolov5} trained on images from the PoTATO dataset~\cite{batista2024potato} with a square input resolution of $1224\times1224$ pixels. At inference, detections are filtered with a confidence threshold of $0.6$ using YOLO’s default non-maximum suppression, and the centroid of each bounding box is used as a potential target point for downstream processing. On the PoTATO test split, the model reaches $\mathrm{mAP}@0.5 = 0.921$. Despite differences between our camera mounting and that in PoTATO, no additional fine-tuning on new images is performed given preliminary experiments presented satisfactory results.

\subsection{Calibration and Projection}
\label{subsec:pp-calib-proj}

\newcommand{\tildev}[1]{\tilde{\mathbf{#1}}}
\newcommand{\bv}[1]{\mathbf{#1}}

In order to establish the relation between image and robot frame coordinates, we first estimate the camera intrinsics. Using checkerboard images spanning the field of view, we calibrate the intrinsic matrix $\mathbf{K}\!\in\!\mathbb{R}^{3\times 3}$ under the pinhole model, and coefficients $\mathbf{d}$ for the Plumb Bob distortion model.

\begin{figure}[htb]
    \centering
    \begin{subfigure}[b]{0.492\linewidth}
        \centering
        \includegraphics[width=\linewidth]{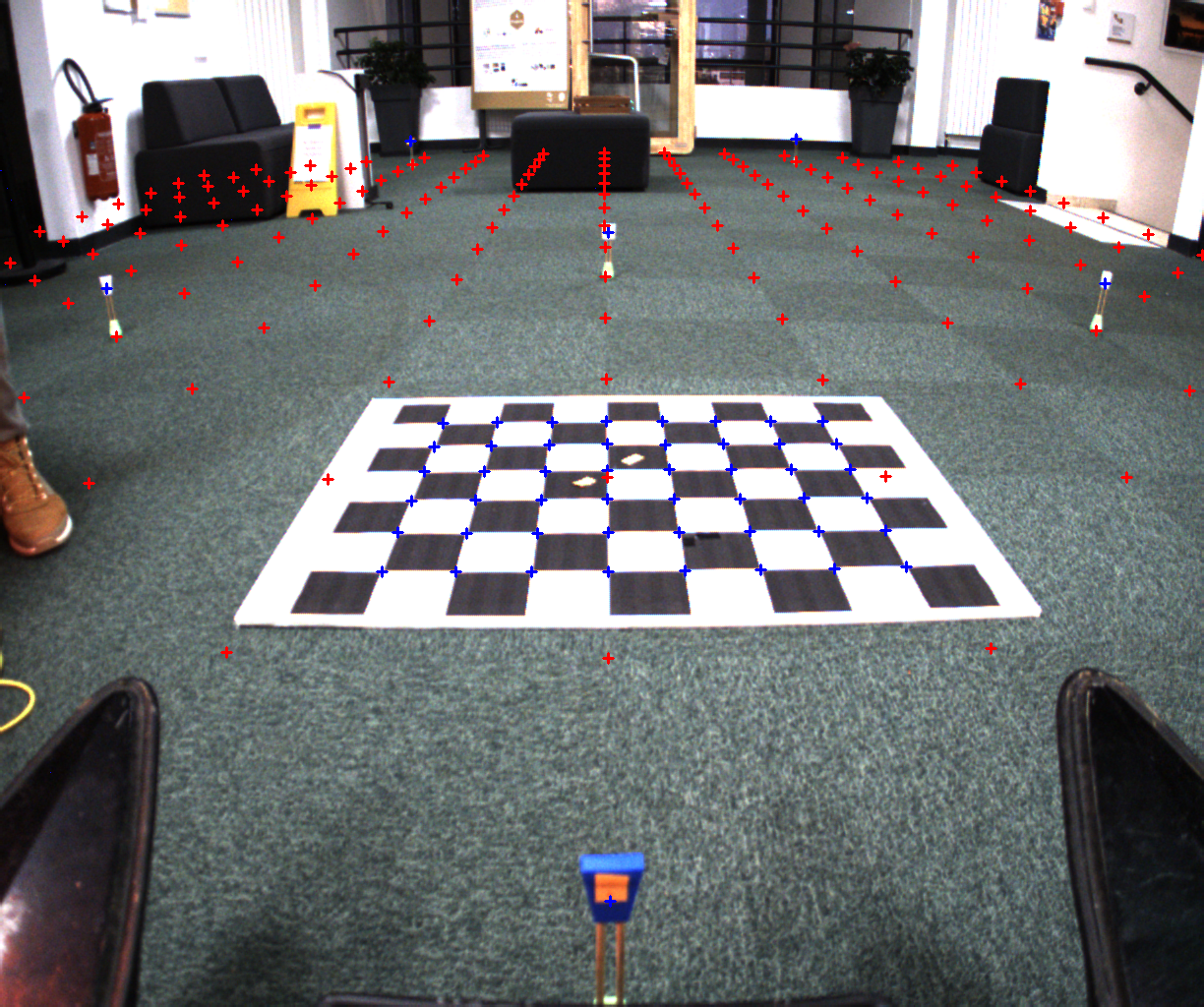}
    \end{subfigure}
    \hfill
    \begin{subfigure}[b]{0.492\linewidth}
        \centering
        \includegraphics[width=\linewidth]{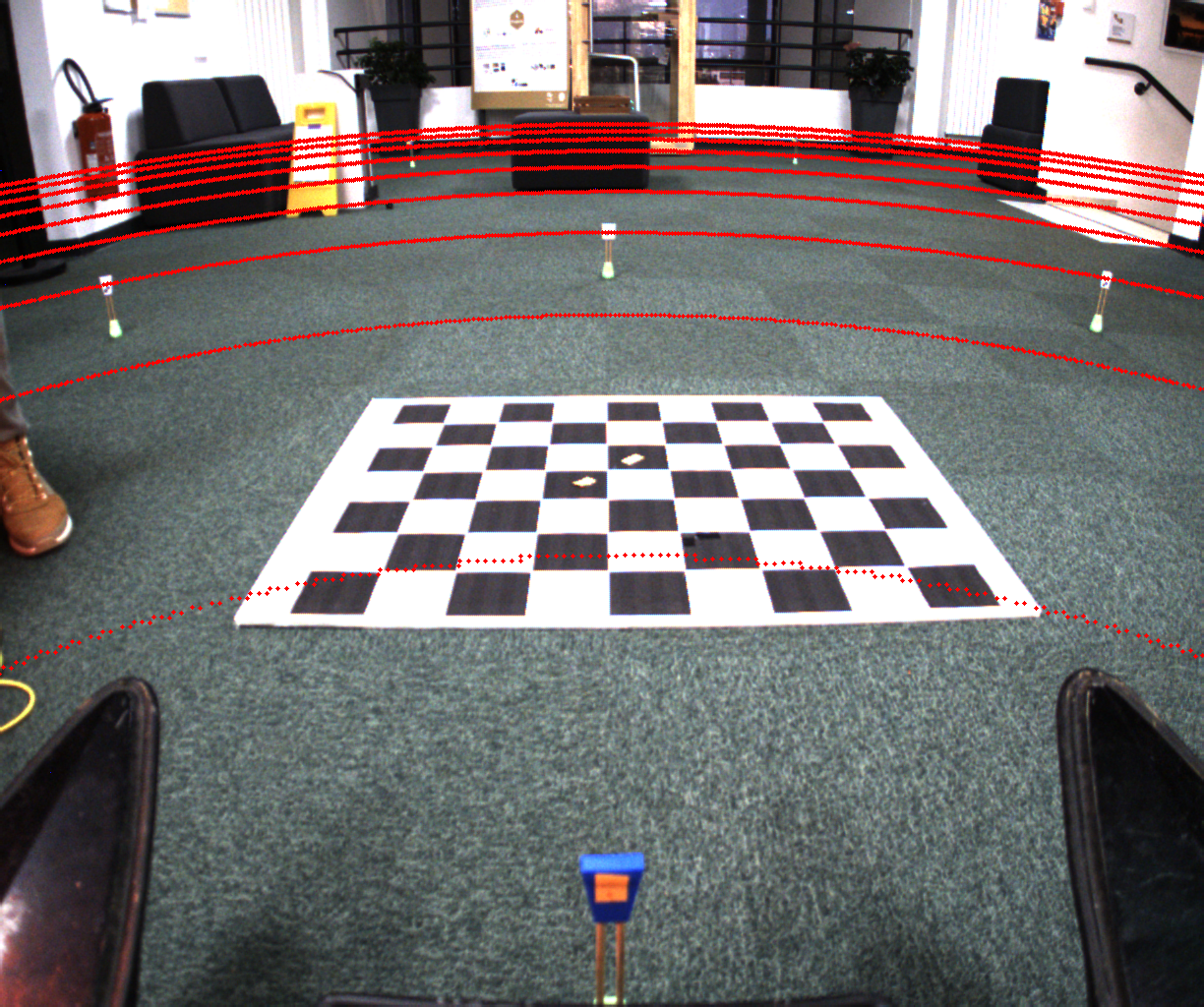}
    \end{subfigure}
    \caption{Calibration and range visualization. \emph{Left}: calibration with fiducial markers used to estimate the camera–robot transform. \emph{Right}: water-plane distance contours (1–9\,m) projected into the image using the calibrated homography.}
    \label{fig:camera-calibration}
\end{figure}

To compute the extrinsics, we place the ASV in a neutral pose consistent with the pitch observed during field trials and distribute fiducials across three distinct planes at known locations in the robot frame (see \cref{fig:camera-calibration}). For each fiducial, we record its robot-frame coordinates and its undistorted pixel location in the calibrated image, thereby establishing a set of 2D--3D correspondences. Applying a standard PnP solver to these correspondences yields the camera pose with respect to the robot frame, parameterized by the rotation $\mathbf{R}_{cr}\!\in\!SO(3)$ and translation $\mathbf{t}_{cr}\!\in\!\mathbb{R}^3$, which are used in the subsequent formulation.

\chg{Because the camera is rigidly mounted to the vessel, this calibration is used as a fixed camera--robot transform at nominal pitch, providing a simple and reproducible real-time projection model without online attitude compensation. Pitch-induced projection error is therefore treated as a known limitation.}

Following projective geometry (see, e.g.,~\cite{szeliski2022cvbook}), the calibrated intrinsics $\mathbf{K}$ and extrinsics $(\mathbf{R}_{cr},\mathbf{t}_{cr})$ can be used to define the pinhole projection matrix $P \;=\; \mathbf{K}\,[\,\mathbf{R}_{cr}\mid \mathbf{t}_{cr}\,],$ and $\mathbf{R}_{cr}=\begin{bmatrix}\bv r_1&\bv r_2&\bv r_3\end{bmatrix}$.
On the water surface we use the plane constraint $Z=0$, removing the $\bv r_3$ term and simplifying the plane-induced homography $H \;=\; K\,[\,\bv r_1\;\;\bv r_2\;\;t_{cr}\,]\in\mathbb{R}^{3\times 3}.$

With $\tildev x_u=[u\;v\;1]^T$ denoting the undistorted image coordinates and $\sim$ the homogeneous equality, we can transform a point $(X,Y)$ from the robot frame to image plane using \cref{eq:fwd-water}, Similarly, given an undistorted pixel position $\tildev x_u$ we can recover water-plane coordinates via \cref{eq:inv-water}.

\begin{equation}
\label{eq:fwd-water}
\tildev x_u \;\sim\; H \begin{bmatrix}X\\Y\\1\end{bmatrix}
\end{equation}

\begin{equation}
\label{eq:inv-water}
\begin{bmatrix}X\\Y\\1\end{bmatrix} \;\sim\; H^{-1}\,\tildev x_u
\end{equation}


\subsection{Perception Pipeline Characterization}
\label{subsec:perception-pipeline}

In this section, we characterize the perception pipeline in real and simulated settings, reporting on-board end-to-end latency, introducing a simulated perception pipeline that mirrors the real camera, and quantifying projection errors.

\subsubsection{Inference and end-to-end latency on the real platform}
We deploy the trained YOLO detector on the Jetson Xavier with TensorRT acceleration and profile the end-to-end latency of the perception stack. The pipeline is decomposed into five stages:
\begin{itemize}
    \item \textit{Capture}: delay from sensor exposure (message timestamp) until the processing node receives the image; this includes Ethernet switch, camera driver, and ROS transport.
    \item \textit{Extraction}: debayering the raw polarimetric mosaic and computing the glare-suppressed image using the equations in \cref{subsec:polarimetric-object-detection}.
    \item \textit{Preprocess}: conversion to the network tensor (resize/normalize) expected by YOLO.
    \item \textit{Detection}: TensorRT-optimized inference of the pretrained network on the Jetson Xavier GPU.
    \item \textit{Projection}: projecting the detections from the image plane to the robot frame via \cref{eq:inv-water}.
\end{itemize}

Empirical timings are summarized in \cref{tab:perception-pipeline-benchmarking}. The \textit{Extraction} stage presents the highest latency ($\sim70\%$). The mean end-to-end latency is $\sim250\ \mathrm{ms}$. In practice we operate at $2\ \mathrm{Hz}$ to preserve headroom and avoid queuing under load. For temporal consistency, each projected goal is published with the original camera sensor timestamp, enabling downstream modules to compensate for the accumulated delay. \chg{Specifically, the goal is expressed in the global frame, assuming a stationary target, and is transformed into the local frame at control time using the current state estimate. Because localization is provided by high-rate, high-accuracy RTK-GPS, this timestamp-consistent transformation compensates for perception latency and minimizes its effect on the control loop, despite the latency in the image processing pipeline}.

\begin{table}[htb]
  \centering
  \small
  \caption{End-to-end perception pipeline latency on a Jetson Xavier: per-stage statistics (ms; mean, std.\ dev., min, max) computed over $\sim$500 images.}
  \label{tab:perception-pipeline-benchmarking}
\begin{tabular}{lrrrr}
\toprule
Step & Mean & Std Dev & Min & Max\\
\midrule
Capture & 16.03 & 1.41 & 15.13 & 32.96 \\
Extraction & 180.00 & 12.73 & 169.87 & 225.34 \\
Preprocess & 11.35 & 1.87 & 9.86 & 27.10 \\
Detection & 42.52 & 0.80 & 41.81 & 52.95 \\
Projection & 0.23 & 0.37 & 0.14 & 7.69 \\
Total Time & 248.09 & 3.73 & 241.90 & 261.49 \\
\bottomrule
\end{tabular}
\end{table}

\subsubsection{Simulated perception pipeline.}
We simulate the perception pipeline by reusing the calibrated projection model from \cref{subsec:pp-calib-proj}, instead of simulating a photorealistic camera running an object detection model. 
Specifically, targets on the water plane $(X,Y,0)$ are projected to the image with \cref{eq:fwd-water} and reprojected back to robot-frame goals with \cref{eq:inv-water}. 
This design mirrors the real ASV perception characteristics while remaining lightweight and deterministic, enabling controlled studies of how perception imperfections affect control. In particular, we can inject controlled errors such as pixel localization noise in the image plane and pitch bias to quantify their impact on goal accuracy and downstream controller performance.

We also use the simulated perception in some field trials on the real ASV by bypassing the detector and generating programmatic goals with the calibrated projection model. This enables isolating and validating control and dynamics under real disturbances while preserving a deterministic perception abstraction that matches simulation.

\subsubsection{Projection error characteristics.}
\begin{figure}[htb]
  \centering
  \includegraphics[width=\linewidth]{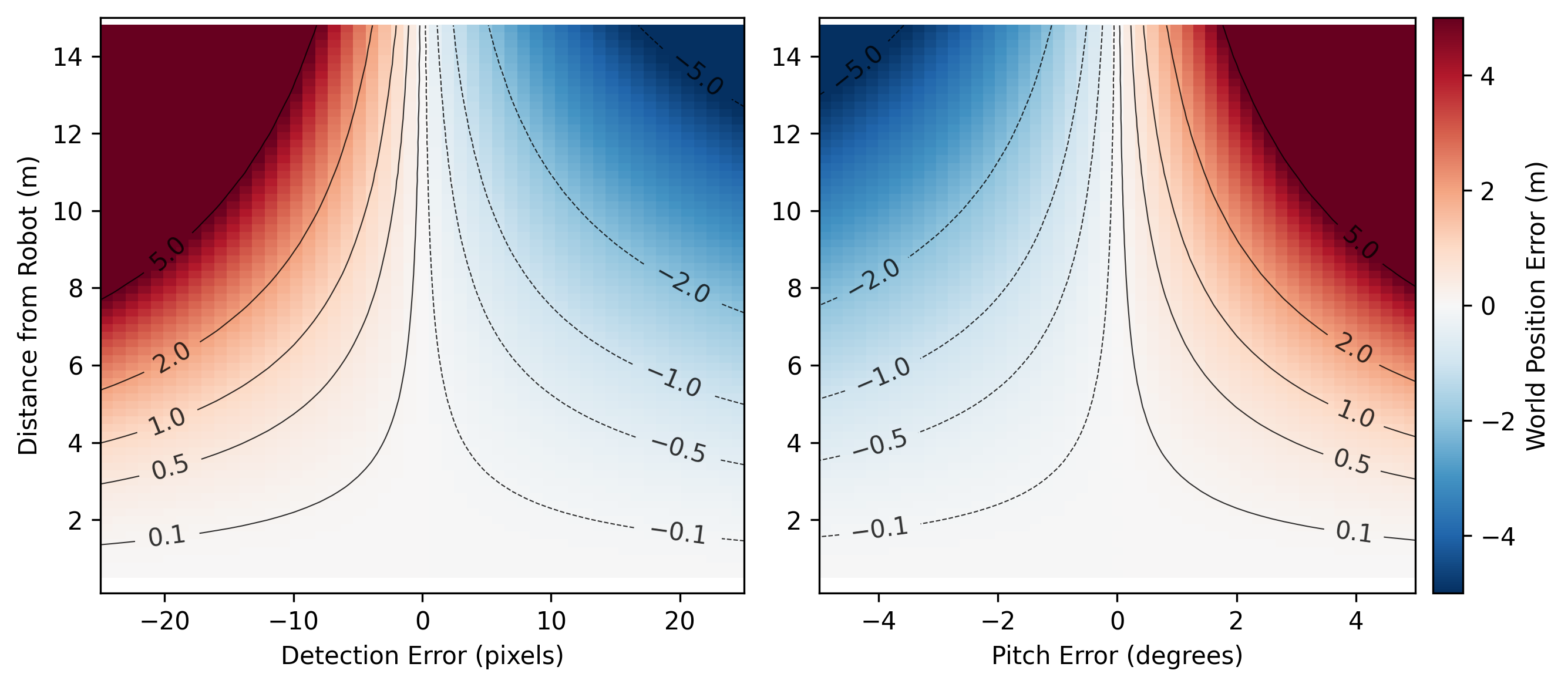}
  \caption{Planar position error vs.\ distance-to-target. Errors are larger at long range and diminish as the ASV approaches the object.}
  \label{fig:proj-error}
\end{figure}

Anticipating the integration with the control component, we provide an analysis of goal detection errors that can arise from the projection model described in \cref{subsec:pp-calib-proj}.
Two dominant sources are considered: (i) image-space localization noise (pixel-level uncertainty on the detection center) and (ii) extrinsic calibration errors (pitch bias relative to the neutral calibration). Using the calibrated intrinsics and extrinsics, we propagate each error source through \cref{eq:inv-water} and report the resulting goal position error as a function of target range. 
The resulting curves (\cref{fig:proj-error}) show that both effects increase with distance; even small image or pitch perturbations produce large errors at long range, and the induced errors decrease rapidly as the ASV closes in, where the homography becomes well conditioned.
Analyzing data from field trials, we observe pitch variation within $\pm 2^{\circ}$ and detection errors within $\pm 10$\,px. For these ranges (\cref{fig:proj-error})
at $r\!\approx\!6$\,m the induced error is on the order of $1$\,m, and within $r\!\le\!2$\,m it falls below $0.1$\,m. Overall, these effects have a much larger impact on distance estimation at long range and become negligible at close range.

In summary, this subsection measured end to end perception delay, defined a lightweight perception abstraction, and characterized projection errors. These results ground controller design, enable latency compensation, and provide a clear basis for interpreting field behavior. The perception abstraction further supports progressive testing using deterministic goals that are repeatable in simulation and field trials, as evaluated in~\cref{sec:control} and ~\cref{sec:integration}.

\section{REINFORCEMENT LEARNING CONTROL}
\label{sec:control}
This section details the implementation of a reinforcement-learning–based low-level controller. A lightweight policy with demonstrated zero-shot simulation-to-real transfer is described, enabling robust point-goal navigation. Field experiments are presented to evaluate performance, identify limitations, and articulate key design considerations.

\subsection{Problem Formulation}
\label{sec:problem}

Our problem is modeled as a Markov Decision Process (MDP;~\cite{puterman2014markov}) described by the quintuple $\langle \mathcal{S}, \mathcal{A}, \mathcal{T}, \mathcal{R}, \gamma \rangle $. Here, $\mathcal{S}$ denotes the state space, $\mathcal{A}$ the action space, $\mathcal{T}(s' \mid s, a)$ the transition function, $\mathcal{R}(s,a)$ a scalar reward, and $\gamma \in [0,1]$ the discount factor. At each discrete time $t$, the agent observes a state $s_t \in \mathcal{S}$, samples an action $a_t \in \mathcal{A}$ from its policy $\pi(a_t \mid s_t)$, receives a reward $\rho_t = \mathcal{R}(s_t, a_t)$, and transitions to the next state $s_{t+1} \sim \mathcal{T}(\cdot \mid s_t, a_t)$. 

Formally, we do not have access to the full state of the system, but instead to observations. However, for simplification, and because our RL agent is based on Proximal Policy Optimization (PPO)~\cite{schulman_proximal_2017}, we consider the observation to be the state ($s_t := o_t$).
To summarize, at every step the environment provides a state observation $s_t$, the reward $\rho_t$, and a termination flag $\delta_t \in \{0,1\}$. The agent’s objective is to maximize the expected discounted return $J(\pi)$ over episodes of horizon $T$.

\begin{equation}
J(\pi) = \mathbb{E}_{\pi}\!\left[ \sum_{t=0}^{T} \gamma^t\, \mathcal{R}(s_t, a_t) \right]
\end{equation}

\subsection{Methodology}
\label{sec:control-methodo}

\subsubsection{Simulation Setup}

The dynamic model of the ASV used in the simulation is based on Fossen’s six degrees of freedom model~\cite{fossen2011handbook}, which governs the motion of the vehicle in response to internal and external forces:
\begin{equation}
    \mathbf{M}\dot\nu + \mathbf{D}(\nu)\nu + g(\eta) = \tau_{\text{thruster}} + \tau_{\text{disturbance}}
    \label{eq:fossen}
\end{equation}
Here, $\eta = [x, y, z, \phi, \theta, \psi]^T$ is the position vector, and $\nu = [u, v, w, p, q, r]^T$ is the velocity vector, composed of surge, sway, heave, roll, pitch, yaw. $\mathbf{M}$ is the system inertia matrix and $\mathbf{D}(\nu)$ is the hydrodynamic damping matrix, given by:
\begin{equation}
    \mathbf{D}(\nu) = \mathbf{D}_l + \mathbf{D}_q(\nu)
    \label{eq:damping}
\end{equation}

It consists of linear ($\mathbf{D}_l$) and quadratic ($\mathbf{D}_q(\nu)$) terms, representing the linear and quadratic hydrodynamic damping coefficients in surge, sway, and rotation. Also in \cref{eq:fossen}, $g$ represents gravitational and buoyancy forces and moments. The terms $\tau_{\text{thruster}}$ and $\tau_{\text{disturbance}}$ refer to forces and moments generated by the thruster system and external disturbances (e.g., wind, waves), respectively. The effects of Coriolis forces are not considered in this model.

\subsubsection{Task Description}

Similarly to the approach described in~\cite{batista2024drl4asv}, the RL agent is based on the PPO algorithm, using an actor-critic architecture with two hidden layers of 64 units each. This setup is designed for efficient control of the ASV’s thrusters in dynamic environments. 

The dynamic model is implemented in simulation to train an agent capable of autonomously controlling the ASV’s thrusters for the task of capturing floating waste. To perform this task, the ASV navigates over the waste, capturing it with a fixed net positioned between its hulls. In simulation, the physics engine simulates the full 3D state, but the control objective and policy operate on the planar manifold because the water surface is modeled as locally flat, enabling a $2D$ formulation.

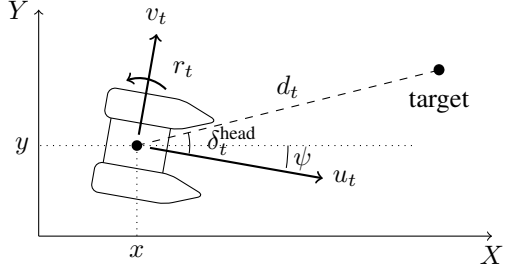
\begin{figure}[htb]
    \centering
    \usetikzlibrary{calc, angles, quotes}

\newcommand{\drawboat}[1][]{
    \begin{scope}[#1]
        \begin{scope}[shift={(-0.5,0)}] 
            \draw[rounded corners=1mm] 
            (-0.2,-0.5) -- (-0.2,0.5) -- (0,1) -- (0.2,0.5) -- (0.2,-0.5) -- cycle;
        \end{scope}
        \begin{scope}[shift={(0.5,0)}] 
            \draw[rounded corners=1mm] 
            (-0.2,-0.5) -- (-0.2,0.5) -- (0,1) -- (0.2,0.5) -- (0.2,-0.5) -- cycle;
        \end{scope}
        \begin{scope}[shift={(0,-0.4)}] 
            \draw[] (-0.3,0) -- (0.3,0);
        \end{scope}
        \begin{scope}[shift={(0,0.4)}] 
            \draw[] (-0.3,0) -- (0.3,0);
        \end{scope}
        \node at (0,0) (base_link) {$.$};
        \fill (0,0) circle (2pt); 

        \draw[->, thick] (base_link) -- ++(0,2.5) node[anchor=west] (u_arrow) {$u_{t}$};
        \draw[->, thick] (base_link) -- ++(-1.5,0) node[anchor=south] {$v_{t}$};
        
        \coordinate (v_tip) at ($(base_link)+(-0.8,0.25)$);
        \draw[->, thick] (v_tip) arc (145:210:0.5) node[pos=0.1, anchor=south west] {$r_{t}$};
    \end{scope}
}

\begin{tikzpicture}[scale=1.0] 
    \draw[->] (0,0) -- (6,0) node[anchor=north] {$X$};
    \draw[->] (0,0) -- (0,3) node[anchor=east] {$Y$};

    \drawboat[xshift=1.3cm, yshift=1.2cm, rotate=-100] 
    \coordinate (base_link) at (1.3,1.2); 
    \coordinate (T) at (5.3,2.2);
    \node at (T) {$\circ$};
    \fill (T) circle (2pt);
    \node[below of=T, node distance=12pt] {target}; 
    
    \draw[dotted] (base_link) -- (base_link |- 0,0) node [at end, below] {$x$};
    \draw[dotted] (base_link) -- (base_link -| 0,0) node [at end, left] {$y$};
    \draw[dotted] (base_link) -- (base_link -| 5,0) node [at end, right] {};
    
    \draw[dashed] (base_link) -- (T);
    \path (base_link) -- (T) node [midway, above] {$d_{t}$};

    \coordinate (u_tip) at ($(base_link)+(2.0,-0.3)$);
    \coordinate (y_ax) at ($(base_link)+(3.0,0,0)$);
    \pic [draw, -, "$\delta_{t}^\text{head}$", angle eccentricity=1.8, angle radius=0.7cm] {angle = u_tip--base_link--T};
    \pic [draw, -, "$\psi$", angle eccentricity=1.1, angle radius=2cm] {angle = u_tip--base_link--y_ax};

\end{tikzpicture}
    \caption{Definition of reference coordinate frame and target position in the simplified $2D$ space used for the task.}
    \label{fig:dynamic_model_diagram}
\end{figure}

As described in \cref{fig:dynamic_model_diagram}, the target is specified in a two-dimensional space. Accordingly, states, goals, and rewards are computed from planar variables; out-of-plane motions (heave, roll, pitch) are treated as exogenous disturbances that affect dynamics but are not part of the policy input or objective.

The coordinates \(x\) and \(y\) represent the position of the ASV’s center in a fixed world frame, and \(d_{t}\) indicates the planar distance to the goal. The task is considered successful when the ASV passes within \(0.15\ \text{m}\) of the target’s position relative to its center. This is a conservative threshold since the boat’s dimensions allow capture of floating waste at distances up to \(0.30\ \text{m}\). To focus solely on evaluating the capture task, the environment is kept free of obstacles. Perception field-of-view constraints are not enforced during training; the goal may be placed at any relative bearing and range with respect to the boat.

\textbf{Action space: }
The action space, $\mathbf{u}_t = [\mathbf{u}_{t}^{\text{left}} \quad \mathbf{u}_{t}^{\text{right}}]^T$, consists of the command inputs for the left and right thrusters, respectively, at time step $t$. These inputs are transformed into thrust forces based on the characteristic force curves obtained from system identification experiments.

\textbf{Observation space: }
At each time step $t$, the observation space is defined as $\mathbf{o}_t \in \mathbb{R}^{8}$:

\begin{equation}
    \mathbf{o}_t = [\textbf{u}_{t-1}^{\text{left}}\  \textbf{u}_{t-1}^{\text{right}}\ u_{t}\ v_{t}\ \chg{r_{t}}\ \cos(\delta_t^{\text{head}})\ \sin(\delta_t^{\text{head}})\ d_{t}]^T
\end{equation}

The components consist of actions from the previous step $\textbf{u}_{t-1}^{\text{left}}$ and $\textbf{u}_{t-1}^{\text{right}}$. The surge speed $u_t$, sway speed $v_t$, and yaw rate $r_t$, which describe the linear and angular velocities of the ASV in the body frame. The orientation of the ASV relative to the goal is represented by $\cos(\delta_t^{\text{head}})$ and $\sin(\delta_t^{\text{head}})$, providing a continuous representation of the bearing of the target. The final element $d_{t}$ denotes the distance to the goal at time step $t$ (\cref{fig:dynamic_model_diagram}).

\textbf{Reward function: }

We design a compound per-step reward $\rho_t$ (\cref{eq:reward}) to encourage progress toward the goal, alignment of the vessel’s heading, smooth and energy-aware actuation, operation within a desired velocity range, time pressure, and a terminal success bonus.
The reward function parameters, detailed in \cref{tab:reward_table_all}, have been meticulously tuned to enhance the agent's performance.

\begin{table*}[tbh]
\centering
\caption{Per-step rewards, weights, and parameters}
\renewcommand{\arraystretch}{1.2}
\setlength{\tabcolsep}{4pt}
\resizebox{\linewidth}{!}{
\begin{tabular}{l l l l p{0.36\linewidth}}
\toprule
\textbf{Term} & \textbf{Equation} & \textbf{Weight} & \textbf{Parameters (value / unit)} & \textbf{Purpose} \\
\midrule
$\rho_{1_t}$
  & $(d_{t-1}-d_t) $
  & $\lambda_{1}=1.0$
  & —
  & Reward distance progress. \\
$\rho_{2_t}$
  & $\big(\cos\delta_{t}^\mathrm{head}-\cos\delta_{t-1}^\mathrm{head}\big) $
  & $\lambda_{2}=5.0$
  & —
  & Reward bearing progress. \\
$\rho_{3_t}$
  & $max\{0,\,1-(\delta_{t}^\mathrm{head}/\delta_{\mathrm{thr}})^2\,\}$
  & $\lambda_{3}=0.01$
  & $\delta_{\mathrm{thr}}=0.1$ rad
  & Bonus for small bearing values. \\
$\rho_{4_t}$
  & $(|\mathbf{u}_{t}^{\mathrm{left}}|^2+|\mathbf{u}_{t}^{\mathrm{right}}|^2)\,{\Delta t}/{E_{\max}}$ 
  & $\lambda_{4}=-0.1$
  & $E_{\max}=2.0$ [–], $\Delta t$ [s]
  & Penalize excessive actuation. \\
$\rho_{5_t}$
  & $e^{\,\kappa\,\max\!\{0,\;v_t-v_{\max},\;v_{\min}-v_t\}}-1$
  & $\lambda_{5}=1.0$
  & $v_{\min}=0$\; $v_{\max}=0.6$\; $\kappa=-10$
  & Penalize operation outside the desired velocity band. \\
$\rho_{6_t}$
  & $1$
  & $\lambda_{6}=-0.05$
  & —
  & Apply time pressure to shorten episode duration. \\
$\rho_{7_t}$
  & $ \begin{cases}1, & d_t < d_{\mathrm{thr}} \\ 0 & \text{otherwise}\end{cases} $
  & $\lambda_{7}=10.0$
  & $d_{\mathrm{thr}}=0.1$ m
  & One-time success reward upon reaching the goal. \\
  
\bottomrule
\end{tabular}}
\label{tab:reward_table_all}
\end{table*}

In \cref{tab:reward_table_all}, $d_t$ is the distance to the goal, $\delta^\mathrm{head}$ is the bearing (heading error) to the goal, $d_{\mathrm{thr}}$ and $\delta_{\mathrm{thr}}$ are distance and bearing thresholds, $\Delta t$ is the step time, $E_{\max}$ normalizes energy, and $v_t$ is the forward body-frame speed constrained by $[v_{\min},\,v_{\max}]$. Hyperparameters $\lambda_{1\ldots 7}$ weight each term, and $\kappa$ sets the velocity-penalty curvature. 
A low value for $d_{\text{thr}}$ was deliberately chosen to be lower than the actual distance required to capture the target. 
The reward function is then:

\begin{equation}
\label{eq:reward}
    \rho_t \;=\; \sum_{n=1}^{7} \lambda_n \rho_{n_t}
\end{equation}

\subsubsection{Policy Training}
\label{sec:policy-training}

\begin{table}[htb]\centering
\caption{Main PPO hyperparameters}
\begin{tabular}{l c}
\toprule
\multicolumn{2}{c}{{Runner / Data Collection}}\\
\midrule
Horizon length (steps) & 48 \\
Max iterations & 200 \\
\midrule
\multicolumn{2}{c}{{Policy / Networks}}\\
\midrule
Actor hidden dims & $64 \times 64$ (ELU) \\
Critic hidden dims & $64 \times 64$ (ELU) \\
Init action std ($\sigma_0$) & 1.0 \\
\midrule
\multicolumn{2}{c}{{Optimization (PPO)}}\\
\midrule
Learning rate & $5\times10^{-4}$ \\
Schedule & adaptive \\
PPO epochs per update & 5 \\
Mini-batches per update & 4 \\
Clip range $\epsilon$ & 0.2 \\
Value loss coef.\ $c_v$ & 1.0 \\
Value loss clipping & Yes \\
Discount $\gamma$ & 0.99 \\
GAE $\lambda$ & 0.95 \\
Target / desired KL & 0.01 \\
\bottomrule
\end{tabular}
\label{tab:rl_config}
\end{table}

Training was carried out on an NVIDIA GeForce RTX~3080 (10\,GB VRAM) and completed in approximately 10 minutes \chg{for a single 200-iterations run}. \cref{tab:rl_config} presents the key learning configurations that were used. \chg{Targets were spawned in polar coordinates relative to the robot, with range $r \in [1,10]\,\mathrm{m}$ and bearing $\phi \in [-\pi,\pi]$, both sampled from a uniform distribution.}  We employed domain randomization comprising: (i) observation noise (sensor noise) with position $\pm 0.03\,\mathrm{m}$ and orientation $\pm 0.025\,\mathrm{rad}$; (ii) actuation noise (action perturbations); (iii) randomized initial forward velocity $v_0 \sim \mathcal{U}(-0.7,\,0.7)\,\mathrm{m/s}$; (iv) center-of-mass (CoM) perturbations uniformly sampled within $\pm 0.05\,\mathrm{m}$ to model payload shifts; and (v) external wrench disturbances emulating wind and currents with forces $F_x,F_y \sim \mathcal{U}(-0.3,\,0.3)\,\mathrm{N}$ and yaw torque $\tau_z \sim \mathcal{U}(-0.3,\,0.3)\,\mathrm{N\!\cdot\! m}$.
We additionally applied up to $10\%$ multiplicative variation in thruster effectiveness to account for battery state and partial loss-of-effectiveness, 
while hydrodynamic parameters were kept fixed \chg{to preserve the identified model and to avoid excessively enlarging the training space.} Training was repeated with five independent random seeds \chg{to assess initialization and exploration robustness. Across seeds} learning curves and final returns converged to comparable values and the learned policies exhibited qualitatively similar behavior\chg{. Since performance was similar across seeds,} a single seed was \chg{arbitrarily} selected for all subsequent experiments and evaluation.

\subsubsection{Evaluation Metrics}

\newcommand{\PsiNT}{\Psi_{\text{NT}}}
\newcommand{\PsiNE}{\Psi_{\text{NE}}}
\newcommand{\PsiER}{\Psi_{\text{ER}}}
\newcommand{\PsiPD}{\Psi_{\text{PD}}}
\newcommand{\PsiFD}{\Psi_{\text{FD}}}
\newcommand{\PsiSR}{\Psi_{\text{SR}}}

\begin{table*}[t]
\caption{Definition of the 6 metrics used to assess policy performance. The table reports the symbol, name, mathematical formulation, unit, and brief interpretation of each metric. $d_1$ and $d_N$ represent initial and final distance respectively. }
\label{tab:metrics-defition}
\centering
\small
\begin{tabularx}{\linewidth}{llclL}
\toprule
\textbf{Metric} & \textbf{Equation} & \textbf{Unit} & \textbf{Type} & \textbf{Description} \\
\midrule
$\PsiNT$ (\textit{Normalized Time}) &
$\displaystyle \PsiNT = \frac{T}{d_{1}}$ &
$\mathrm{s}\,\mathrm{m}^{-1}$ &
\textit{efficiency} &
Episode duration normalized by straight-line distance to goal. \\
$\PsiNE$ (\textit{Normalized Energy}) &
$\displaystyle \PsiNE = \frac{\sum_{t=1}^{N}\!\big(\lvert \mathbf{u}_{t}^{\mathrm{left}}\rvert + \lvert \mathbf{u}_{t}^{\mathrm{right}}\rvert\big)}{d_1}$ &
$\text{(au)}\mathrm{m}^{-1}$  &
\textit{efficiency} &
Accumulated actuator effort normalized by straight-line distance to goal. \\

$\PsiER$ (\textit{Excess Rotation}) &
$\displaystyle \PsiER = \sum_{t=2}^{N}\lvert \psi_{t} - \psi_{t-1}\rvert - \lvert \delta_{1}^{\mathrm{head}}\rvert$ &
$\mathrm{rad}$ &
\textit{stability} &
Total heading change beyond the initial bearing-to-goal offset. \\

$\PsiFD$ (\textit{Final Distance}) &
$\displaystyle \PsiFD = d_N$ &
$\mathrm{m}$ &
\textit{accuracy} &
Terminal distance to goal. \\

$\PsiPD$ (\textit{Path Deviation}) &
$\displaystyle \PsiPD = \sum_{t=2}^{N}\lvert d_{t} - d_{t-1}\rvert + d_{N} - d_{1}$ &
$\mathrm{m}$ &
\textit{accuracy} &
Excess path length relative to a straight-line distance to goal. \\

$\PsiSR$ (\textit{Success Rate}) &
$\displaystyle
\PsiSR =
\begin{cases}
1, & d_{N} \le 0.15~\mathrm{m} \\
0, & \text{otherwise}
\end{cases}$ &
— &
\textit{task completion} &
Binary success under a $0.15$\,m tolerance. \\

\bottomrule
\end{tabularx}
\end{table*}

One of the key focus of our work is evaluating the robustness of the deployed RL policy and the sim-to-real transferability. 

We test the robustness of the RL policy with respect to different types of disturbances using complementary metrics to access efficiency, stability, accuracy, and task completion. Although trials may continue until a minimum-distance criterion is met, metrics are computed solely on the trajectory of the first approach to the goal, thereby excluding any post-pass corrections; this choice aligns with the application and prevents metric distortion. The first approach is defined as the first crossing of the line through the goal that is orthogonal to the start–goal segment (see \cref{fig:prog-goal-failure}). All data recorded after this crossing are excluded from metric computation, and the ASV’s position at the crossing is used to compute the \emph{success rate}. The same procedure is applied consistently in simulation and field experiments. The definitions, units, and interpretations of all metrics are summarized in \cref{tab:metrics-defition}.

We also evaluate our policies using a sim-to-real gap metric based on~\cite{da_survey_2025}. This allows to evaluate transferability and prioritize improvements that might be needed to increase simulation fidelity.

Let $\mathcal{E}_{sim}$ be the simulated environment and $\mathcal{E}_{real}$ the real world.  As introduced in \cref{sec:sota-robustness},
the policy trained in simulation solves the MDP $\mathcal{M}_s:\langle \mathcal{S}_s,\mathcal{A}_s,\mathcal{T}_s,r_s, \gamma \rangle$, corresponding to $\mathcal{E}_{sim}$. However, when deployed in $\mathcal{E}_{real}$, the MDP $\mathcal{M}_r$ is different.

Let $\pi_s$ be the policy the RL agent learned from $\mathcal{M}_s$. Let $\Psi$ be the evaluation metric selected to quantify the performance of the policy: $\Psi_s$ corresponds to the evaluation conducted in $\mathcal{E}_{sim}$, and $\Psi_r$ in $\mathcal{E}_{real}$.
Over $n$ experiments, the sim-to-real gap $\mathcal{G}(\pi)$ can be formulated as:


\begin{equation}
\label{eq:sim2real-gap}
\mathcal{G}(\pi)=\frac{1}{n}\sum_{i=1}^{n}
\left|\Psi_s^{i}(\pi_s)-\Psi_r^{i}(\pi_s)\right|
\end{equation}

$\Psi$ is measured identically in $\mathcal{E}_{sim}$ and $\mathcal{E}_{real}$. $\Psi$ may represent any task-relevant evaluation metric. In this work, we use $\PsiNT$, $\PsiNE$, $\PsiER$, $\PsiPD$, $\PsiFD$, $\PsiSR$, described in \cref{tab:metrics-defition}.

\subsection{Experiments}
To assess policy robustness, we conducted matched experiments in simulation and in field trials (\cref{fig:diagram-test}). Specifically, we evaluate the zero-shot robustness of a policy trained only in simulation, and we characterize how different types of perturbations affect performance under realistic conditions. Experiments in this section use the simulated perception (SP) introduced in \cref{fig:diagram-perception} and formalized in \cref{subsec:perception-pipeline}. The real perception (RP) pipeline is used in \cref{sec:integration}.

For evaluation, each trial consisted of a sequence of predefined goal positions at distances \(3, 6, 9\,\mathrm{m}\) and bearings spanning \([-30^\circ, 30^\circ]\) in \(15^\circ\) increments, matching the camera’s field of view and covering representative goal configurations. The identical goal set was repeated under 14 experimental conditions to quantify performance degradation under distinct disturbance types. Using the perception abstraction, we reproduced the same goal sequences and disturbance profiles in both simulation and field trials, achieving experiment parity and isolating the control policy from perception and planning effects. This setup allows injecting controlled disturbances while holding perception constant, enabling direct measurement of control performance degradation and assessment of the sim-to-real gap.

\chg{
The 14 disturbance conditions considered in this study are summarized in \cref{tab:exp_conditions} and reflect representative challenges of ASV-based floating-waste collection. Among them, \textit{03-NoRateLim} and \textit{04-LDandNRL} specifically evaluate a modeling discrepancy identified in preliminary experiments: the \textit{Kingfisher}'s microcontroller enforces a command rate limit that is not represented in the nominal simulator, leading to systematic differences between commanded and realized thrust (\cref{subsec:hardware-platform}, \cref{fig:cmd-drive-response}). To isolate the effect of this mismatch, we additionally trained a policy under nominal settings, differing only in that it omits the empirically identified actuator rate limiter. This policy is used only in \textit{03-NoRateLim} and \textit{04-LDandNRL}. All remaining experiments use the policy trained using the improved thruster model.
}

\begin{table*}[tbh]
\color{black}
\caption{\chg{Summary of the 14 experimental conditions with different types of disturbances}}
\label{tab:exp_conditions}
\centering
\scriptsize
\setlength{\tabcolsep}{4pt}
\renewcommand{\arraystretch}{1.12}
\begin{tabularx}{\textwidth}{l l X}
\toprule
\textbf{Title} & \textbf{Disturbance type} & \textbf{Description} \\
\midrule
\textit{01-Ideal} & None & Baseline with nominal conditions and no deliberate perturbations. \\
\midrule
\textit{02-LocDelay} & States & A fixed \(100\,\mathrm{ms}\) delay is injected into the localization estimate before it is consumed by the policy. \\
\textit{03-NoRateLim} & Actions & Policy trained in simulation without the actuator rate limiter enforced on the MCU, creating a deliberate sim-to-real mismatch. \\
\textit{04-LDandNRL} & States + Actions & Combination of \textit{02-LocDelay} and \textit{03-NoRateLim}, emulating mis-modeled actuation and mis-tuned localization. This correctable case is excluded from other combined settings. \\
\midrule
\textit{05-ThrLoE10} & Actions & Loss of efficiency on the right thruster with command scaling \(u^{right}=(1-\alpha)u^{right}\), using \(\alpha=0.10\), corresponding to \(10\%\) loss of efficiency. \\
\textit{06-ThrLoE30} & Actions & Same as \textit{05-ThrLoE10}, with \(\alpha=0.30\), corresponding to \(30\%\) loss of efficiency. \\
\textit{07-ThrLoE50} & Actions & Same as \textit{05-ThrLoE10}, with \(\alpha=0.50\), corresponding to \(50\%\) loss of efficiency. \\
\midrule
\textit{08-DynPert} & Transitions & Moderate dynamic perturbation. In simulation, the total mass is increased from \(35\) to \(41.25\,\mathrm{kg}\), the lateral center of gravity is shifted to \(\mathrm{cog}_y=-0.10\,\mathrm{m}\), and the quadratic surge damping coefficient is increased from \(17.26\) to \(25.89\). In field trials, a \(5.2\,\mathrm{kg}\) payload is attached with a \(10\,\mathrm{cm}\) rightward offset from the platform center (see \cref{fig:kingfisher-net}). \\
\textit{09-DynPertStr} & Transitions & Same as \textit{08-DynPert}, with \(\mathrm{cog}_y=-0.20\,\mathrm{m}\), quadratic surge damping increased to \(34.52\), and the \(5.2\,\mathrm{kg}\) payload attached with a \(20\,\mathrm{cm}\) rightward offset from the platform center. \\
\midrule
\textit{10-PNoise05px} & States & Image-plane perception noise with perturbation radius \(r=5\) pixels. The target projection is perturbed in the image plane and reprojected to the robot frame. Noise is resampled for each new frame (\(2\,\mathrm{Hz}\)). \\
\textit{11-PNoise25px} & States & Same as \textit{10-PNoise05px}, with \(r=25\) pixels. \\
\textit{12-PNoise50px} & States & Same as \textit{10-PNoise05px}, with \(r=50\) pixels. \\
\midrule
\textit{13-CombPert} & Combined & Moderate combined perturbation comprising \textit{05-ThrLoE10}, \textit{08-DynPert}, and \textit{10-PNoise05px}. \\
\textit{14-CombPertStr} & Combined & Strong combined perturbation comprising \textit{06-ThrLoE30}, \textit{09-DynPertStr}, and \textit{12-PNoise50px}. \\
\bottomrule
\end{tabularx}
\end{table*}

\chg{
\subsection{RL-based Control Results and Discussion}
\label{subsec:rl-control-results}
}

\begin{table*}[!t]
\centering
\caption{Mean performance $(\Psi)$ and sim-to-real gap $(\mathcal{G})$ metrics for each experimental condition. For each metric/column, the values with the highest degradation are highlighted.}
{
\setlength{\tabcolsep}{5pt} 
\scriptsize
\begin{tabular}{@{}l*{6}{c}|*{6}{c}|*{6}{c}@{}}
\toprule
\textbf{Experiment} &
\multicolumn{6}{c}{\textbf{Simulation}} &
\multicolumn{6}{c}{\textbf{Field}} &
\multicolumn{6}{c}{\textbf{Sim-to-real Gap}} \\
\cmidrule(lr){2-7}\cmidrule(lr){8-13}\cmidrule(l){14-19}
 & $\PsiNT$ & $\PsiNE$ & $\PsiFD$ & $\PsiER$ & $\PsiPD$ & $\PsiSR$
 & $\PsiNT$ & $\PsiNE$ & $\PsiFD$ & $\PsiER$ & $\PsiPD$ & $\PsiSR$
 & $\mathcal{G}_\text{NT}$ & $\mathcal{G}_\text{NE}$ & $\mathcal{G}_\text{FD}$ & $\mathcal{G}_\text{ER}$ & $\mathcal{G}_\text{PD}$ & $\mathcal{G}_\text{SR}$ \\
\midrule
\textbf{01-Ideal} &1.9 &32.52 &0.02 &0.21 &0.11 &1 
&1.89 &32.88 &0.04 &0.18 &0.13 &1 
&0.01 &0.36 &0.02 &0.03 &0.02 &0 \\
\textbf{02-LocDelay} &1.87 &31.34 &0.04 &0.31 &0.12 &1 
&1.88 &32.92 &0.04 &0.21 &0.14 &1 
&0.01 &1.58 &0 &0.1 &0.02 &0 \\
\textbf{03-NoRateLim} &\textbf{2.12} &\textbf{58.51} &\textbf{0.07} &\textbf{2.22} &\textbf{0.22} &\textbf{0.93}
&\textbf{2.27} &\textbf{64.35} &\textbf{0.12} &\textbf{2.15} &\textbf{0.35} &\textbf{0.73}
&\textbf{0.15} &\textbf{5.84} &\textbf{0.05} &0.07 &\textbf{0.13} &\textbf{0.20} \\
\textbf{04-LDandNRL} &\textbf{2.16} &\textbf{60.41} &\textbf{0.12} &\textbf{2.31} &\textbf{0.31} &\textbf{0.67}
&\textbf{2.23} &\textbf{63.95} &\textbf{0.15} &\textbf{2.06} &\textbf{0.36} &\textbf{0.6} 
&0.07 &3.54 &0.03 &\textbf{0.25} &0.05 &0.07 \\
\textbf{05-RThrLoE10} &1.92 &34.63 &0.02 &0.19 &0.11 &1
&1.92 &34.8 &0.04 &0.22 &0.13 &1 
&0 &0.17 &0.02 &0.03 &0.02 &0 \\
\textbf{06-RThrLoE20} &2 &40.55 &0.03 &0.21 &0.12 &1 
&1.98 &40.16 &0.05 &0.23 &0.15 &1 
&0.02 &0.39 &0.02 &0.02 &0.03 &0 \\
\textbf{07-RThrLoE50} &2.2 &52.16 &0.03 &0.31 &0.13 &1 
&2.23 &\textbf{53.08} &0.08 &0.32 &0.19 &\textbf{0.93}
&0.03 &0.92 &0.05 &0.01 &0.06 &0.07 \\
\textbf{08-DynPert} &1.94 &35.58 &0.02 &0.25 &0.1 &1 
&2.01 &40.97 &0.04 &0.28 &0.14 &1
&0.07 &5.39 &0.02 &0.03 &0.04 &0 \\
\textbf{09-DynPertStr} &2 &39.14 &0.04 &0.34 &0.15 &1 
&2.04 &42.09 &0.05 &0.25 &0.14 &1 
&0.04 &2.95 &0.01 &0.09 &0.01 &0 \\
\textbf{10-PNoise05px} &1.94 &32.77 &0.02 &0.28 &0.13 &1
&1.92 &34.04 &0.04 &0.2 &0.14 &1 
&0.02 &1.27 &0.02 &0.08 &0.01 &0 \\
\textbf{11-PNoise25px} &1.92 &32.74 &0.03 &0.38 &0.12 &1
&1.93 &35.08 &0.04 &0.29 &0.15 &1 
&0.01 &2.34 &0.01 &0.09 &0.03 &0 \\
\textbf{12-PNoise50px} &1.93 &32.54 &0.03 &0.5 &0.13 &1 
&1.92 &33.55 &0.04 &0.36 &0.14 &1 
&0.01 &1.01 &0.01 &0.14 &0.01 &0 \\
\textbf{13-CombPert} &2 &38.61 &0.03 &0.33 &0.13 &1 
&2.08 &45.59 &0.04 &0.33 &0.13 &1 
&0.08 &\textbf{6.98} &0.01 &0 &0 &0 \\
\textbf{14-CombPertStr} &\textbf{2.35} &\textbf{57.07} &0.04 &\textbf{1.04} &0.17 &1 
&\textbf{2.25} &\textbf{54.91} &0.07 &\textbf{0.63} &0.19 &1 
&\textbf{0.10} &2.16 &0.03 &\textbf{0.41} &0.02 &0 \\
\bottomrule
\end{tabular}
}
\label{tab:metrics-results}
\end{table*}

\begin{figure*}[!t]
    \centering
    \begin{subfigure}[b]{0.48\textwidth}
        \centering
        \includegraphics[width=\linewidth]{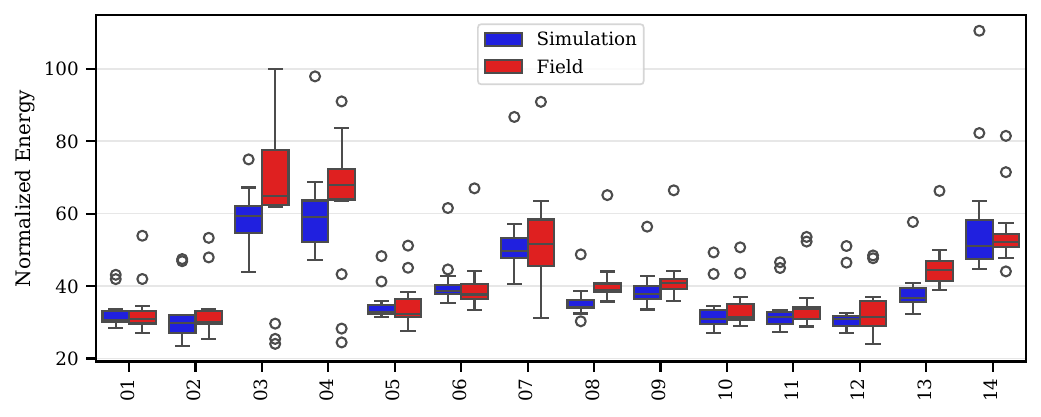}
    \end{subfigure}
    \hfill
    \begin{subfigure}[b]{0.48\textwidth}
        \centering
        \includegraphics[width=\linewidth]{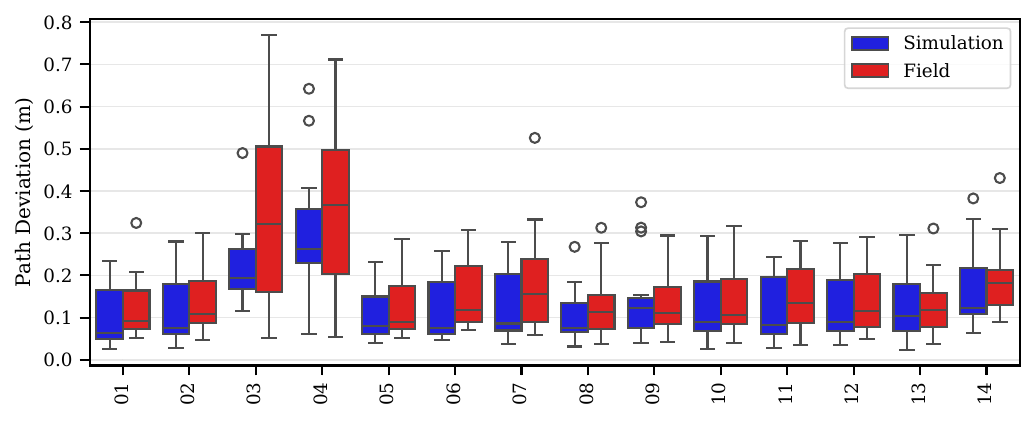}
    \end{subfigure}
    \begin{subfigure}[b]{0.48\textwidth}
        \centering
        \includegraphics[width=\linewidth]{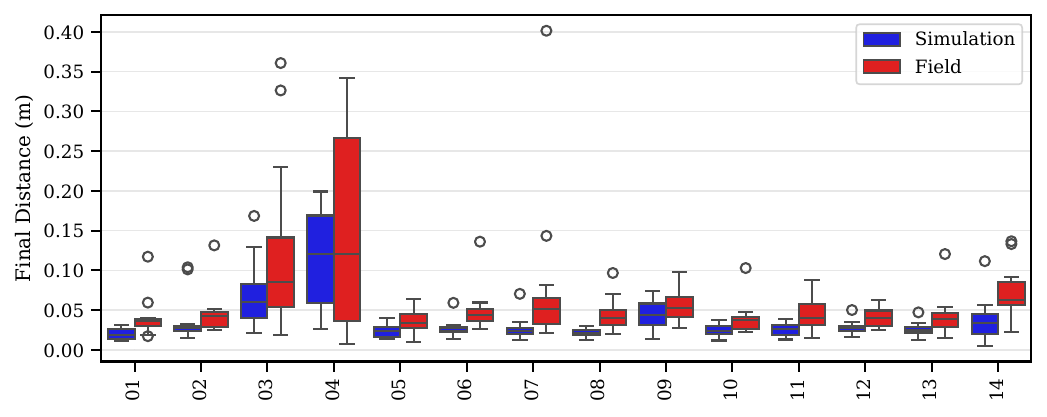}
    \end{subfigure}
    \hfill
    \begin{subfigure}[b]{0.48\textwidth}
        \centering
        \includegraphics[width=\linewidth]{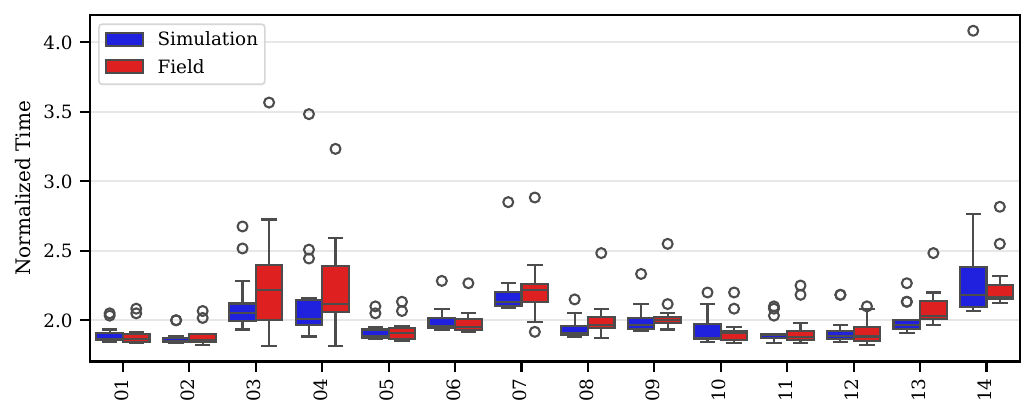}
    \end{subfigure}
    \begin{subfigure}[b]{0.48\textwidth}
        \centering
        \includegraphics[width=\linewidth]{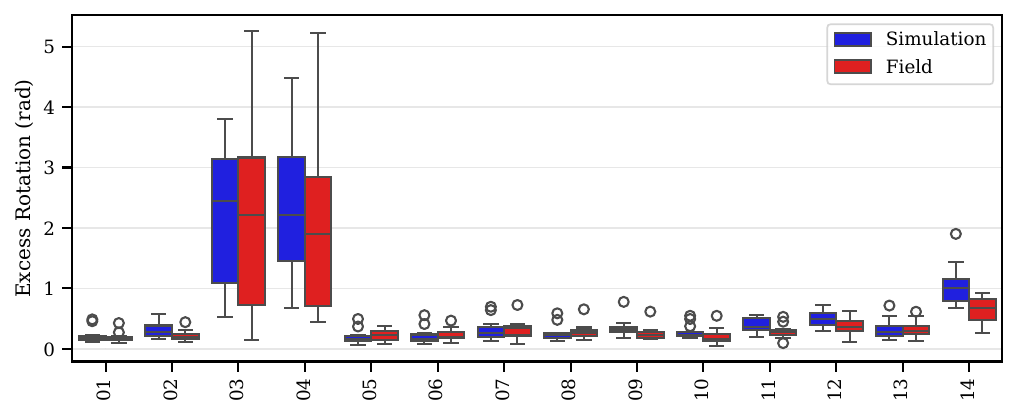}
    \end{subfigure}
    \hfill
    \begin{subfigure}[b]{0.48\textwidth}
        \centering
        \includegraphics[width=\linewidth]{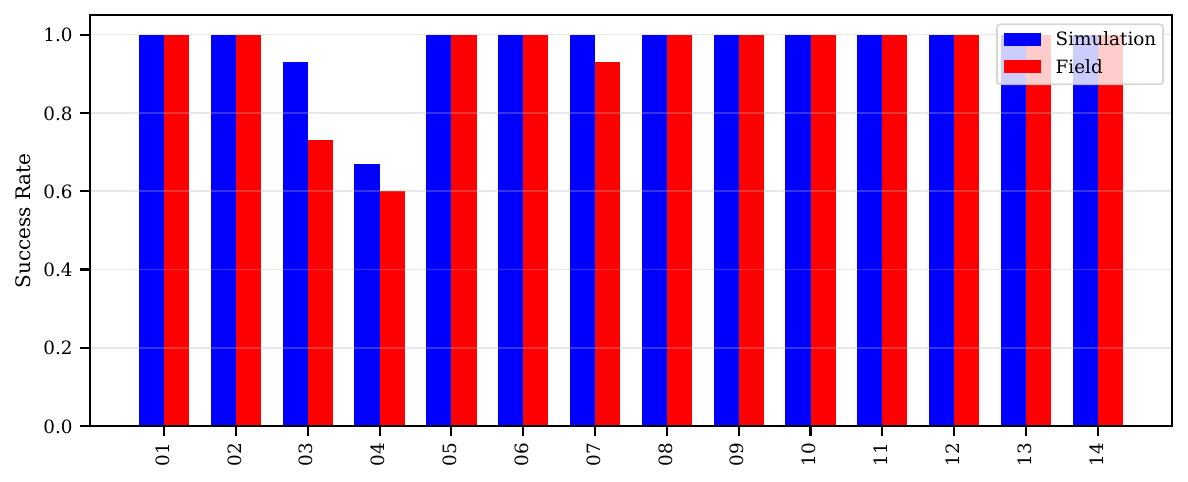}
    \end{subfigure}
    \caption{Bar plots of the six metrics ($\PsiNE, \PsiPD, \PsiFD, \PsiNT, \PsiER, \PsiSR$)  across all experimental conditions. Each bar plot corresponds to metrics from the 15 goals shown in \cref{fig:prog-goal-traj}. Degradation is pronounced for \textit{03-NoRateLim} and \textit{04-LDandNRL} due to mismodeled thruster dynamics; other disturbances show robust performance with only modest variation. Results are largely consistent between simulation and field tests.}
    \label{fig:prog-goal-metrics}
\end{figure*}

\begin{figure*}[tb]
    \centering
        \includegraphics[width=\linewidth]{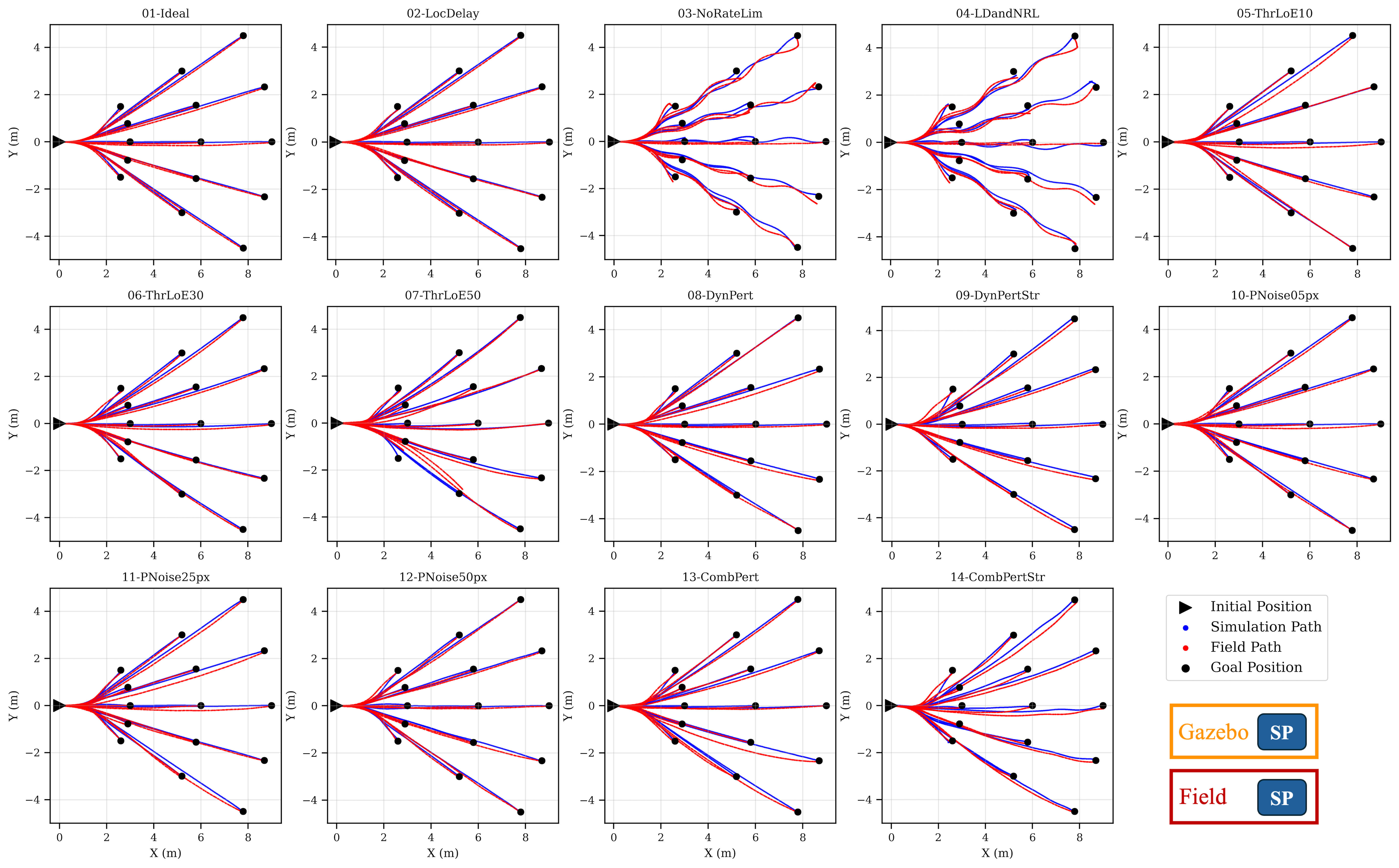}
    \caption{Simulated and field-test paths for all experiments. Most cases show no visible degradation; \textit{03-NoRateLim} and \textit{04-LDandNRL} exhibit rotation-induced oscillations, and \textit{14-CombPertStr} shows a slightly larger deviation from the optimal path. Trajectories after the first crossing of the goal’s orthogonal line are not displayed. Looking closely at \textit{07-ThrLoE50}, one field trial did not reach the minimal distance to the goal (near position $(2,-2)$).}
    \label{fig:prog-goal-traj}
\end{figure*}

Across 14 experimental conditions, each evaluated on 15 programmatic goals, we conducted \chg{$210$ simulation trials and a corresponding set of $210$ matched field trials}. Quantitative performance is \chg {presented across six metrics in \cref{fig:prog-goal-metrics} and summarized in terms of the corresponding mean values and sim-to-real gaps in \cref{tab:metrics-results}.} Qualitative behavior is shown by the actual trajectories in \cref{fig:prog-goal-traj}.

\paragraph{Overall analysis}
These results show that the RL-based controller achieves high \emph{Success Rate} ($\Psi_{SR}$) across a wide range of perturbation scenarios. The success rate stays at $1.0$ in 23 out of 28 experiments in simulation and field trials (\cref{tab:metrics-results}), with the only exceptions being the experiments \textit{03-NoRateLim} and  \textit{04-LDandNRL}, with large disturbance in the Actions due to a mismatch of the actuators models from training and deployment worlds, as well as a single missed approach in \textit{07-ThrLoE50} field tests, an experiment also with high disturbance on the Actions.

\paragraph{Degradations with significant effect on robustness}
A strong degradation is observed in the \textit{03-NoRateLim} condition, which does not include actuator rate limiting during training. In this case, \emph{Excess Rotation} ($\Psi_{ER}$) rises sharply from nominal values around $0.2\,\mathrm{rad}$ (\textit{01-Ideal}) to more than $2.2\,\mathrm{rad}$ in simulation and $2.15\,\mathrm{rad}$ in field (\cref{tab:metrics-results}). This also doubles the \emph{Normalized Energy} ($\Psi_{NE}$), from $\approx 32$ to above $60$. The mismatch highlights the sensitivity of the policy to actuator dynamics, particularly rate limits, which can be difficult to model precisely in proprietary designs and can be confounded by the high inertia of aquatic platforms. 

Similarly, \textit{04-LDandNRL} (localization delay + missing actuator rate limit modeling) shows pronounced degradation across all metrics, with trajectory-level oscillations visible in \cref{fig:prog-goal-traj}. By contrast, \textit{02-LocDelay} alone remains near nominal. Taken together, these results indicate that (i) only modest localization delay, hence small error on the States, has no significant impact, whereas (ii) missing actuator rate limiting, leading to a large error on the actions, especially when combined with localization delays, drives the strongest degradation in efficiency, stability, and success.

\begin{figure}[tbh]
    \centering
    \includegraphics[width=1\linewidth]{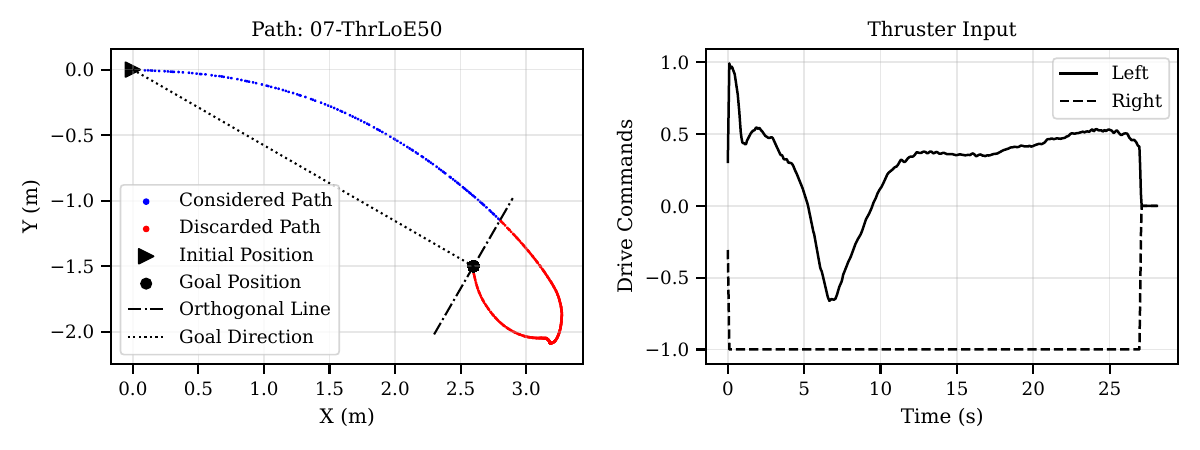}
    \caption{Trajectory and actuator commands showing a missed first approach and recovery under \textit{07-ThrLoE50}. Sustained saturation of the right-thruster command at $-1$ limits clockwise rotation authority, causing the initial miss; the vehicle then reorients and finally reaches the target. For metrics computation, only samples up to the first crossing of the goal-orthogonal line are considered.}
    \label{fig:prog-goal-failure}
\end{figure}

\paragraph{Degradations with limited effect on robustness}
Beyond rate-limit effects, the only other failure occurs under \textit{07-ThrLoE50} (50\% loss-of-effectiveness on the right thruster, high perturbation on the Actions). It fails to reach the goal on the first pass, so $\Psi_{SR}$ drops to $0.93$ (field). Because LoE is applied to the input command signal, the backward force is limited to  $1\,\mathrm{N}$ (\cref{fig:thruster-force-curve}), reducing differential thrust and preventing a sharp clockwise rotation.
Even in these degraded situation, the policy managed to eventually recover and reach the goal, demonstrating that even if not approaching in an optimal distance, recovery was possible. This is presented in \cref{fig:prog-goal-failure}.

For moderate thruster loss-of-effectiveness (\textit{05-ThrLoE10}, \textit{06-ThrLoE30}, smaller perturbations on the Actions), the controller still succeeds in all trials with minimal accuracy loss ($\Psi_{FD} \leq 0.05\,\mathrm{m}$), but efficiency is compromised. $\Psi_{NE}$ increases by $20$--$25\%$ compared to the ideal case, reflecting the additional actuation needed to compensate for reduced thrust. A similar effect is seen in \textit{08-DynPert} and \textit{09-DynPertStr}, perturbations on the Transition, where field values of $\Psi_{NE}$ increase from $32.9$ under ideal conditions to $41.0$ and $42.1$, respectively, while $\Psi_{NT}$ and $\Psi_{FD}$ remain close to nominal. These results suggest a clear robustness--efficiency trade-off: the agent maintains success and accuracy at the cost of higher energy expenditure.

A noteworthy outcome is the resilience to perception noise (i.e State perturbation). Even with severe perturbations of $50$ pixels (\textit{12-PNoise50px}), which exceed by an order of magnitude the noise levels observed in practice (\cref{sec:polarimetric-perception}), the controller achieves $\Psi_{SR}=1.0$ with negligible loss of accuracy ($\Psi_{FD} \leq 0.04\,\mathrm{m}$). Only a small increase in $\Psi_{ER}$ is observed (field: $0.36\,\mathrm{rad}$, compared to $0.18\,\mathrm{rad}$ in the ideal case), indicating strong policy robustness to perception noise.

The combined perturbation experiments further supports the robustness of the trained policy. Under \textit{13-CombPert}, performance remains close to nominal ($\Psi_{FD}=0.04\,\mathrm{m}$, $\Psi_{SR}=1.0$), with an increase in $\Psi_{NE}$ ($45.6$ field vs. $32.9$ ideal). For the stronger combined disturbances \textit{14-CombPertStr}, the degradation becomes more evident: $\Psi_{NT}$ grows to $2.25$ (field), $\Psi_{NE}$ rises close to $55$, and $\Psi_{ER}$ increases to $0.63$. Despite these effects, the success rate remains at $1.0$, and the trajectories in \cref{fig:prog-goal-traj} still converge reliably to the goal.

\paragraph{Sim-to-real gap analysis}
\label{parag:sim-to-real}
The sim-to-real gap (\cref{tab:metrics-results}) is remarkably small in most conditions, typically $\leq 0.05$ for $\Psi_{NT}$ and $\leq 0.02\,\mathrm{m}$ for $\Psi_{FD}$. The largest discrepancies occur in the dynamic perturbation cases (\textit{08}--\textit{09}), where $\Psi_{NE}$ increases by more than $5$ between simulation and field, reflecting the unmodeled hydrodynamic drag and center-of-mass shifts introduced by physical payloads. Nonetheless, the consistently low mean terminal error ($\approx 0.04\,\mathrm{m}$ across field tests) validates the training setup and the effectiveness of domain randomization for zero-shot sim-to-real transfer.

\paragraph{Main insights}

Overall, the dominant vulnerability is \emph{actuator dynamics}: omitting the MCU’s rate limiting during training induces large heading oscillations, substantially increases effort, and reduces success rate. With measured rate limiting and basic domain randomization, the lightweight PPO policy transfers zero-shot, maintaining near-perfect success rate and centimeter-level final distance; degradation appears primarily as an efficiency cost (higher energy and longer time), especially under added mass/drag.
Observation noise in goal detection has limited effect on accuracy and success, producing at most small increases in excess rotation. The only first-pass miss occurs under severe right-thruster loss-of-effectiveness that pushes the system beyond practical control limits (\textit{07-ThrLoE50}). Sim-to-real gaps remain small for time, distance, and success, with the mild discrepancies in normalized energy under dynamic-perturbation loads. For point-goal ASV navigation, simulator fidelity should prioritize accurate actuation modeling and, secondarily, hydrodynamic effects; additional realism in goal-detection noise offers diminishing returns.

\section{INTEGRATION}
\label{sec:integration}

In this section we present an integrated perception-and-control framework for detecting floating waste and autonomously navigating the ASV to collect it. \chg{Such integration is important because a practical autonomous system must combine detection, decision-making, and motion execution in closed loop. In real deployments, this top-level layer could be extended with capabilities such as drift-aware search strategies, mission-level planning, and multi-agent coordination. In this work, however, we deliberately adopt a simple integration strategy, as our primary objective is to validate the interaction between the perception module and the RL-based controller at the system level and how they operate in conjunction rather than to propose a fully developed top-level controller.}  We state the problem, formalize the interfaces, and describe the integration, including a simple goal selection policy that closes the loop between the perception and control modules. The details are provided for clarity and reproducibility, but real applications are encouraged to use more comprehensive approaches. Finally, we present and discuss simulation and field-trial results that validate the joint system performance.

\subsection{Problem Formulation}

The goal is to collect all floating bottles within a known bounded planar region using a single ASV.
The ASV moves on a planar surface with pose \(x_t := (p_t,\theta_t) \in \mathbb{R}^2 \times S^{1}\).
At time \(t\), two finite sets of 2-D points are available, both defined in the global coordinate frame:
the remaining search waypoints \(\mathcal{W}_t \subseteq \mathbb{R}^2\) and the current perception detections \(\mathcal{Z}_t \subseteq \mathbb{R}^2\).
The desired behavior is that the ASV alternates between (i) \emph{Target Search}: traverse the waypoint set to cover the area; and (ii) \emph{Point-Goal Navigation}: when detections are available (\(\mathcal{Z}_t \neq \emptyset\)), temporarily preempt the search to navigate to a selected detection.
After visiting a goal position, resume the search.

To achieve this behavior, we need to design an online goal-selection policy $\mu$ that, given $(x_t,\mathcal{W}_t,\mathcal{Z}_t)$, produces the next goal $g_t=\mu(x_t,\mathcal{Z}_t, \mathcal{W}_t)\in\ \mathcal{W}_t \cup\mathcal{Z}_t$.

\subsection{Methodology}

We implement an integrated perception--control pipeline composed of three core modules: (i) \emph{Perception}, which detects floating waste and outputs detections $\mathcal{Z}_t$ (2-D points on the water plane); (ii) \emph{Control}, which tracks a point goal $g_t$ using a learned RL policy; and (iii) \emph{Integration}, which arbitrates between the remaining coverage waypoints $\mathcal{W}_t$ and the current detections $\mathcal{Z}_t$ to produce the next point goal $g_t\in\mathcal{W}_t\cup\mathcal{Z}_t$. The integration layer operates as a lightweight goal publisher that outputs a single selected point goal.

\textbf{Perception:}
The perception stack uses either the object detection model with real images or the simulated pipeline implementation described in \cref{sec:polarimetric-perception}. The detections on the image plane are projected onto the water plane to yield goal candidates $z \in \mathbb{R}^2$. A short-lived set $\mathcal{Z}_t$ of valid detections is maintained via timeout.

\textbf{Control:}
Low-level actuation is handled by the RL-based controller described in \cref{sec:control}. Given the current state observation and a goal $g_t \in \mathbb{R}^2$, the policy controls the ASV by outputting drive commands for each thruster. 

\textbf{Integration:}
The integration module arbitrates between coverage waypoints and online detections and, at each control cycle, publishes a single point goal for the controller.
At time $t$, let $k_t$ be the index of the next coverage waypoint and $\mathcal{W}_t=(w_{k_t},\dots,w_N)$. Let $\mathcal{Z}_t$ be the set of \emph{active} detections (2-D water-plane points) with timestamps $t_z$ such that $t - t_z \le \tau_{\max}$. A detection is removed upon collection when $\|p_t - z\| \le r_z$, and the current waypoint is marked visited (advancing the index) when $\|p_t - w_{k_t}\| \le r_w$.
To keep detours bounded, preemption is restricted to the \emph{eligible} detections
$\mathcal{E}_t \triangleq \{\, z \in \mathcal{Z}_t \;:\; \|z - w_{k_t}\| \le r_d \,\}$. 
The module then publishes one goal per cycle:
\[
g_t =
\begin{cases}
\displaystyle\arg\min_{z \in \mathcal{E}_t} \|p_t - z\|, & \text{if } \mathcal{E}_t \neq \emptyset,\\[6pt]
w_{k_t}, & \text{otherwise}.
\end{cases}
\]
In summary, the integration module pursues the nearest eligible detection when one is available; otherwise continue along the waypoint tour. The integration module is purely reactive and memoryless. Navigation towards the goal is handled by the RL controller that will steer towards the desired position until a new goal is received. 


\subsection{Experiments}
\label{subsec:integration-experimetns}

We evaluate the integrated system \chg{through three complementary experiment groups designed to provide broad coverage, controlled sim-to-real comparison, and validation with real perception in the loop. \textbf{\emph{E1-Sim-Feasibility}} uses simulation to assess feasibility across a large set of targets. \textbf{\emph{E2-Sim-to-Real}} uses simulated bottle detections to enable direct comparison between simulation and field trials. \textbf{\emph{E3-Real-Perception}} evaluates the integrated system with camera-based real bottle detections in the loop.} Two obstacle-free rectangular test areas were \chg{defined for these experiments}: \emph{Small} ($5\times10\,\mathrm{m}$) and \emph{Large} ($15\times30\,\mathrm{m}$). Each area was discretized into lawnmower waypoint pattern with $5\,\mathrm{m}$ spacing (yielding $6$ and $24$ waypoints, respectively) \chg{to define the waypoint locations used throughout the simulation and field evaluations.}

\chg{\textbf{\emph{E1-Sim-Feasibility}.}}
This experiment stress-tests the arbitration logic between coverage waypoints and detected targets while using abstracted perception. A set of $100$ point targets is randomly sampled within the boundaries of the \emph{Large} test area. The agent must interleave \chg{traversal of the coverage waypoints} with target interception. The experiment is \chg{conducted exclusively} in simulation to assess the feasibility of the \textit{arbitration logic under controlled dynamics}.

\chg{\textbf{\emph{E2-Sim-to-Real}.}}
\chg{This experiment leverages the abstracted perception} to evaluate sim-to-real consistency. A fixed set of target \chg{positions} is executed once in Gazebo and twice in the field on both \emph{Small} and \emph{Large} areas, using identical waypoint sequences and fixed goal locations. The objective is to \chg{compare simulation and field performance for the same programmed trials, thereby characterizing the sim-to-real gap introduced by} real actuation, localization noise, and platform dynamics.

\textbf{\emph{E3-Real-Perception}.}
This experiment validates end-to-end performance under real noisy observations. The full \chg{camera-based} perception pipeline provides target positions used by the RL-based controller, while natural drift caused by wind and water current can generate slow target displacements. \chg{In both the \emph{Small} and \emph{Large} areas, up to $5$ plastic bottles are manually deployed simultaneously at diverse locations. After all bottles are captured, a remote-control (RC) takeover is used to return to shore for unloading, after which autonomous operation resumes with the bottles redeployed as needed until the experiment is completed.}

\chg{Experiments are executed without} artificial disturbances in simulation. \chg{Field tests were conducted within a single day, spanning morning and afternoon sessions under different illumination conditions but similar weather.} To quantify the wind and water current disturbances, we recorded the RTK IMU measurements while letting the ASV drift without any actuation for over 7 minutes. The ASV reached angular speeds between $-0.15$ to $0.08$ rad/s, surge speeds from $-0.14$ to $0.29$ m/s, and sway speeds in the range of $-0.15$ to $0.06$ m/s. Such disturbances are not negligible considering the ASV’s operational velocity of $0.5$ m/s. Additionally, the ASV is also exposed to payload variations when bottles are trapped in the net. Missions are started and stopped manually to accommodate bottle placement and unloading. RC takeover is also used when reaching bottles that drifted beyond the predefined region is necessary.

\chg{
\subsection{Integration Results and Discussion}
}
\label{subsec:integration-results}

In this subsection, we report the results for the three \chg{experiment groups}: \textbf{\emph{E1-Sim-Feasibility}}, \textbf{\emph{E2-Sim-to-Real}}, and \textbf{\emph{E3-Real-Perception}}. \chg{\Cref{tab:integration-metrcis} summarizes the experimental conditions and run-level statistics, including the perception type, grid size, number of bottles, manual intervention, execution time, and total traveled distance. Next, we discuss the main observations for each experiment group and present trajectory plots for the tests in the \emph{Large} area.}

\begin{table*}[htp]
\centering
\color{black}
\scriptsize
\caption{\chg{Run-level statistics for the three integration experiment groups}}
\label{tab:integration-metrcis}
\begin{tabular}{lcccccccrrrr}\toprule
&\multicolumn{3}{c}{\textbf{Environment Details}} & & &\multicolumn{2}{c}{\textbf{Capture}} &\multicolumn{2}{c}{\textbf{Time (s)}} & \\\cmidrule{2-4}\cmidrule{7-10}
\textbf{Experiment Group} &\textbf{Type} &\textbf{Perception Type} &\textbf{Grid Size} &\textbf{Run} &\textbf{Bottles} &\textbf{Autonomous} &\textbf{Manual} &\textbf{Autonomous} &\textbf{Manual} &\textbf{Total Distance (m)} \\\midrule
\textbf{\emph{E1-Sim-Feasibility}} &Gazebo &Simulated (SP) &Large &1 &100 &100 &0 &1509.2 &0 &768.9 \\
\midrule
\multirow{6}{*}{\textbf{\emph{E2-Sim-to-Real}}} &Gazebo &Simulated (SP) &Small &1 &5 &5 &0 &140.4 &0 &73.4 \\
&Gazebo &Simulated (SP) &Large &1 &5 &5 &0 &597.2 &0 &322.5 \\
&Field &Simulated (SP) &Small &1 &5 &5 &0 &142.4 &58.2 &83.8 \\
&Field &Simulated (SP) &Small &2 &5 &5 &0 &133.6 &23.7 &72.8 \\
&Field &Simulated (SP) &Large &1 &5 &5 &0 &609.1 &10.9 &329.4 \\
&Field &Simulated (SP) &Large &2 &5 &5 &0 &589.2 &6.8 &315.5 \\
\midrule
\multirow{5}{*}{\textbf{\emph{E3-Real-Perception}} } &Field &Real (RP) &Small &1 &5 &5 &0 &408.4 &68.5 &198.8 \\
&Field &Real (RP) &Small &2 &15 &13 &2 &1034.7 &123.1 &513.6 \\
&Field &Real (RP) &Large &1 &10 &9 &1 &946.8 &151.2 &502.8 \\
&Field &Real (RP) &Large &2 &5 &4 &1 &938.6 &40 &482.5 \\
&Field &Real (RP) &Large &3 &5 &4 &1 &617.7 &49.8 &335.4 \\
\bottomrule
\end{tabular}
\end{table*}

\subsubsection{{\emph{E1-Sim-Feasibility}}}

\begin{figure}[htb]
    \begin{subfigure}[b]{1.0\linewidth}
        \centering
        \includegraphics[width=\linewidth]{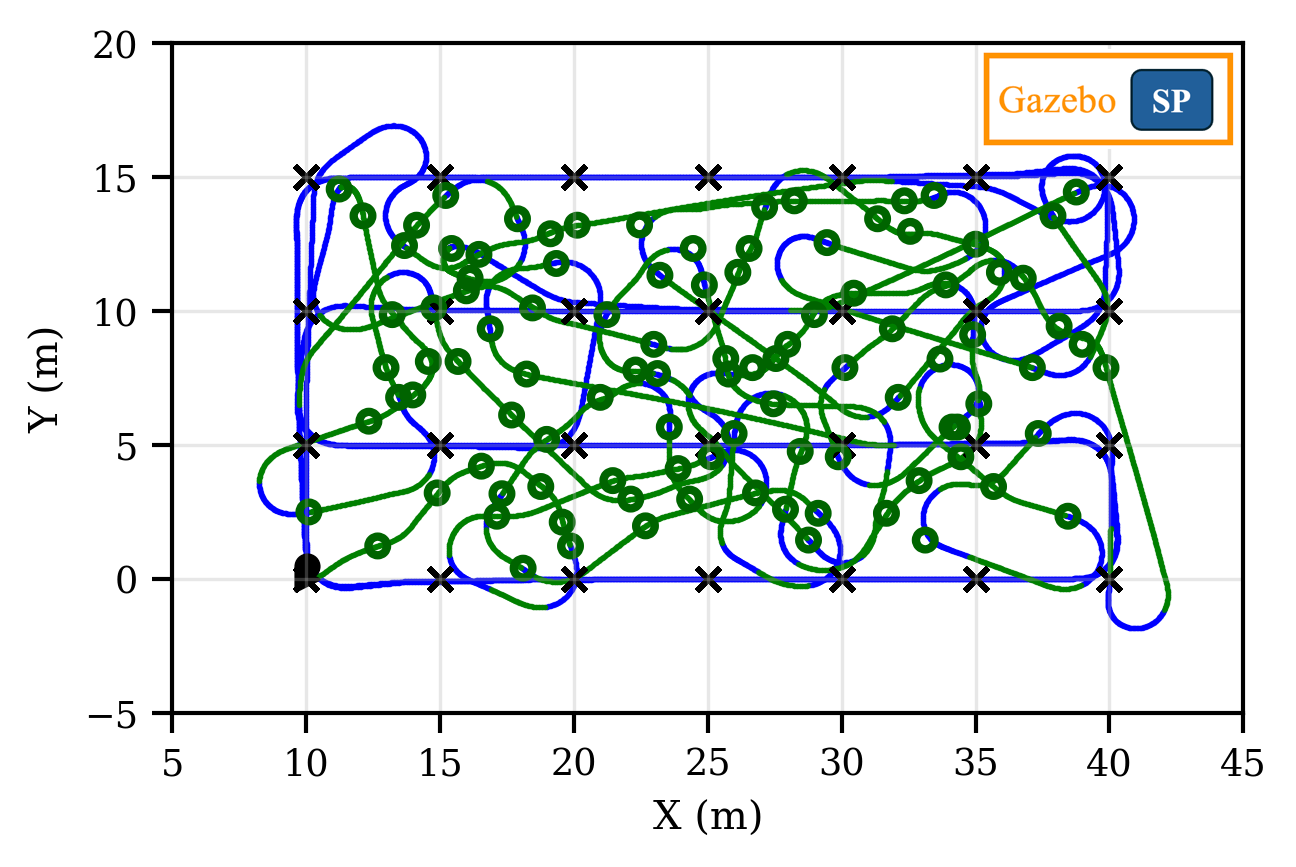}
    \end{subfigure}
    \vspace{0.25em}
    \begin{subfigure}[b]{1.0\linewidth}
        \centering
        \includegraphics[width=\linewidth]{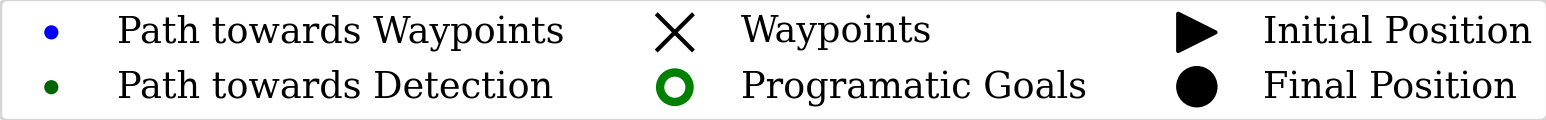}
\end{subfigure}
    \caption{\emph{E1-Sim-Feasibility}: Trajectory on the \emph{Large} area with $100$ random goals. All simulated goals are captured, indicating the feasibility of the proposed goal selection policy.
    }
    \label{fig:waypoint-dense}
\end{figure}

\cref{fig:waypoint-dense} shows the trajectory for $100$ uniformly sampled goals placed over the \emph{Large} area. The agent interleaves \chg{coverage of} the waypoints with detected goals. \chg{The experiment runs for approximately 25 minutes and all targets are captured}. While the alternation policy is naive, these trials demonstrate that the search-and-capture loop is feasible with stable convergence under disturbance-free dynamics. \chg{It also show that the goal selection policy remains effective even when many candidate targets are simultaneously visible in the workspace.} While efficiency and path optimization are left for improved arbitration in future work, this experiment also highlights the value of simulation tests that enable quick large-scale validation that would be challenging to repeat in field trials.

\begin{figure*}[htb]
    \centering
    \begin{subfigure}[b]{0.32\textwidth}\centering
        \includegraphics[width=\linewidth]{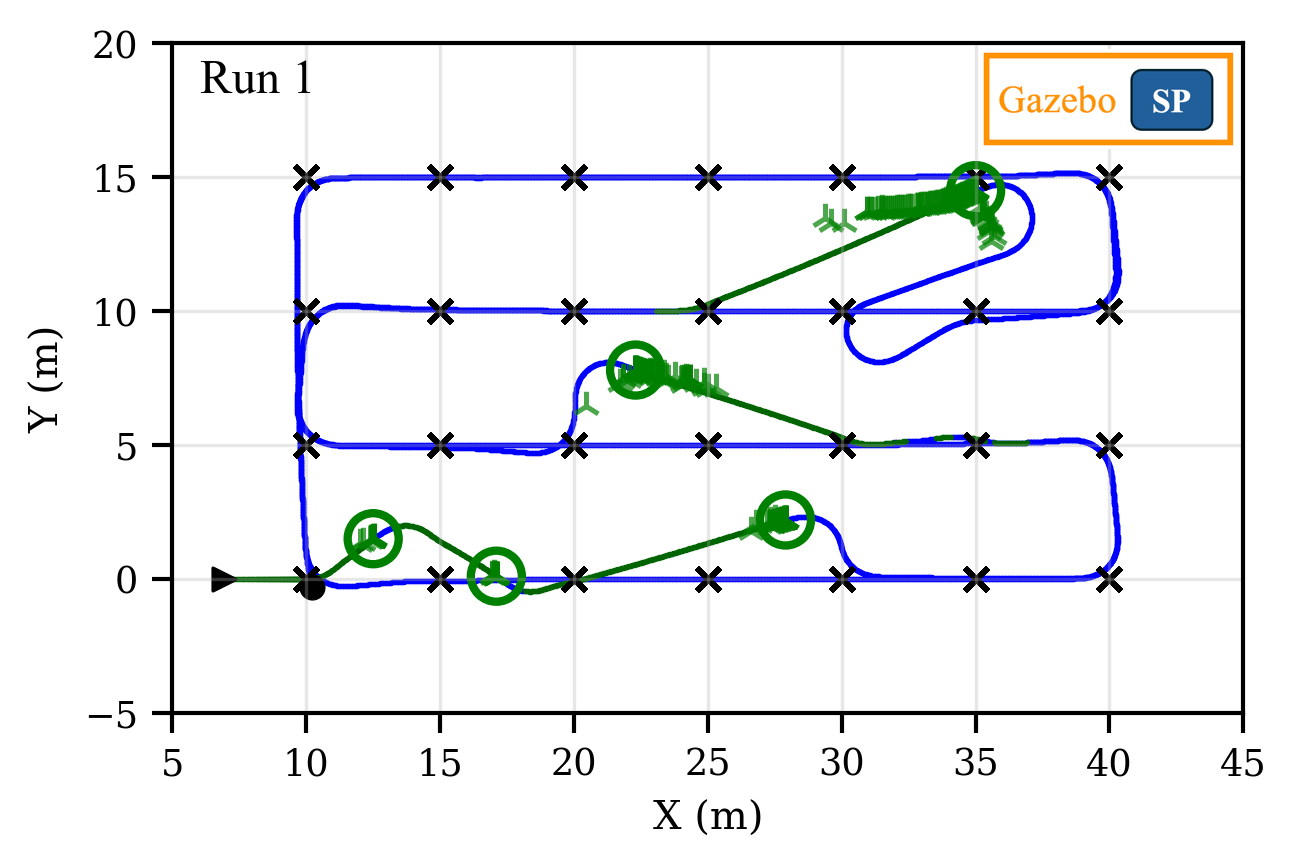}
    \end{subfigure}\hfill
    \begin{subfigure}[b]{0.32\textwidth}\centering
        \includegraphics[width=\linewidth]{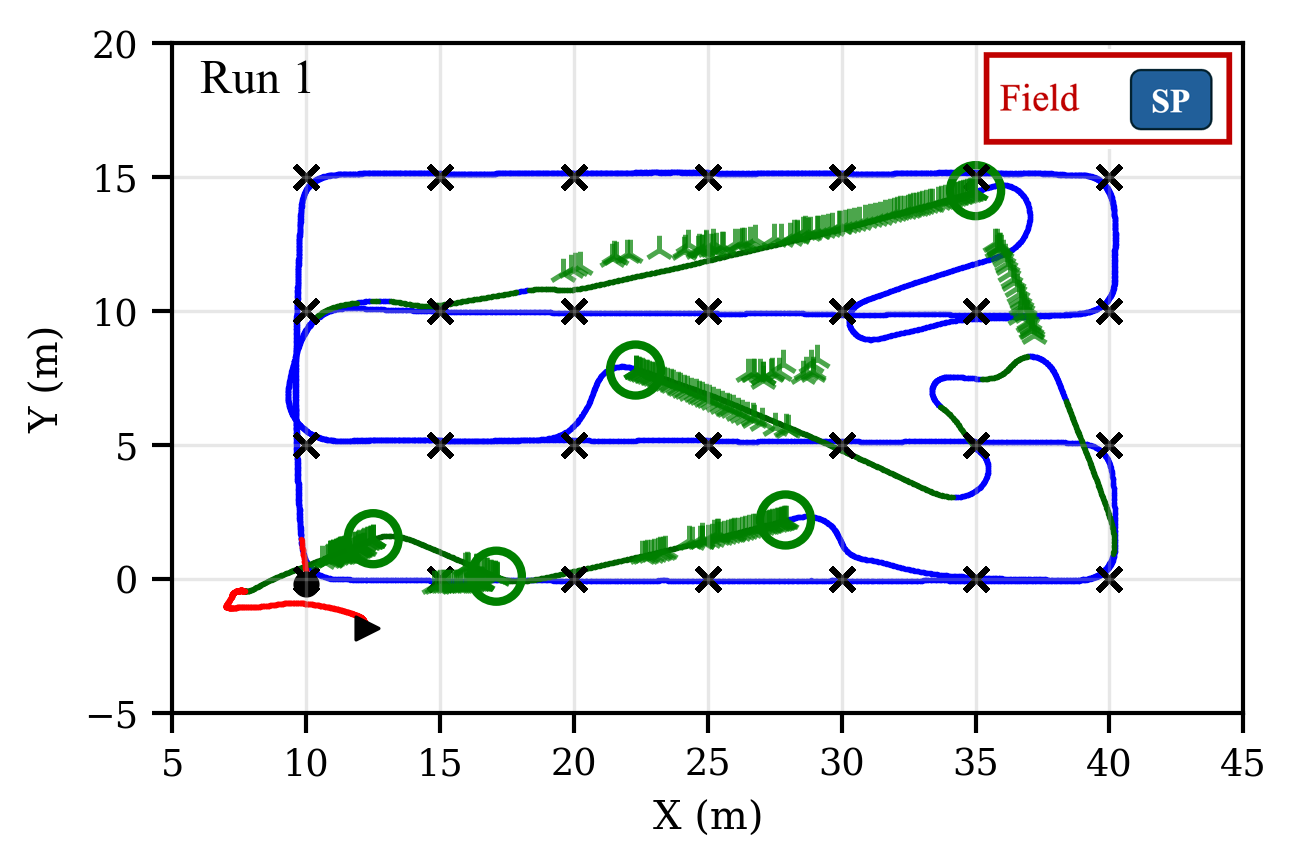}
    \end{subfigure}\hfill
    \begin{subfigure}[b]{0.32\textwidth}\centering
        \includegraphics[width=\linewidth]{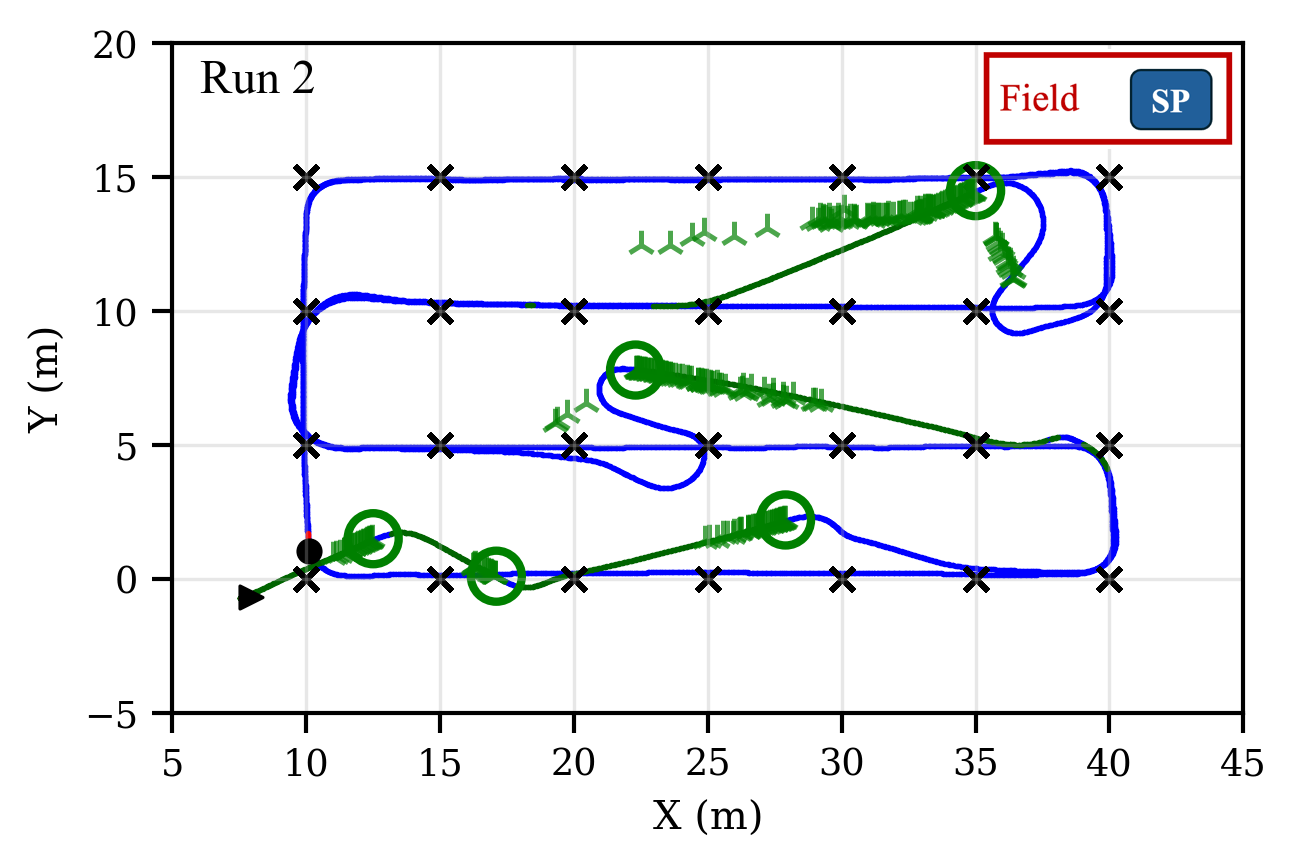}
    \end{subfigure}
    \vspace{0.25em}
    \begin{subfigure}[b]{1.0\textwidth}\centering
        \includegraphics[width=0.92\linewidth]{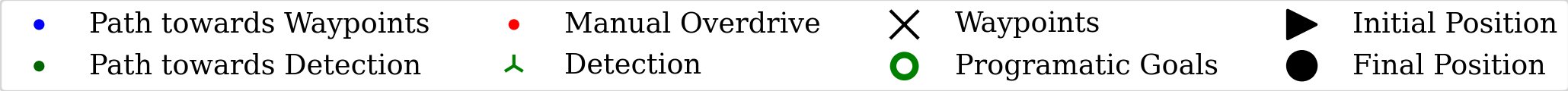}
    \end{subfigure}
    \caption{\emph{E2-Sim-to-Real}: Programmed-goal trajectories across simulation and two field runs. \chg{The trajectories indicate that projection errors can induce transient goal switches and, consequently, trajectory deviations.}}
    \label{fig:waypoints-sim-goals}
\end{figure*}

\subsubsection{\emph{E2-Sim-to-Real} (Programmed Goals)}
\cref{fig:waypoints-sim-goals} compares programmed-goal executions in simulation and across two field repetitions. As a reminder, the programmed-goal experiment \chg{relies on simulated perception}, which projects detections using the calibrated camera parameters (see \cref{subsec:pp-calib-proj}) and is sensitive to pitch-induced projection errors. \chg{Detection \textit{trails} are visible} near each programmed goal, indicating \chg{projection uncertainty}. Detections may occasionally enter the acceptance radius, causing transient goal switches until the integration logic reverts to the waypoints once the distance threshold \(r_d\) rejects the target. This effect is particularly evident in the second field run (\cref{fig:waypoints-sim-goals}), indicating that pitch variations can affect goal selection and lead to trajectory deviations. Despite these transients, all programmed goals are eventually reached autonomously in both Gazebo and field trials. \chg{\Cref{tab:integration-metrcis} reports comparable travel distances and completion times for corresponding grid sizes in simulation and field experiments. The reported manual operation time accounts only for positioning the ASV at the beginning and end of each experiment.}

\subsubsection{\emph{E3-Real-Perception} (Bottles-in-the-Loop)}
\begin{figure*}[htb]
    \centering
    \begin{subfigure}[b]{0.32\textwidth}\centering
        \includegraphics[width=\linewidth]{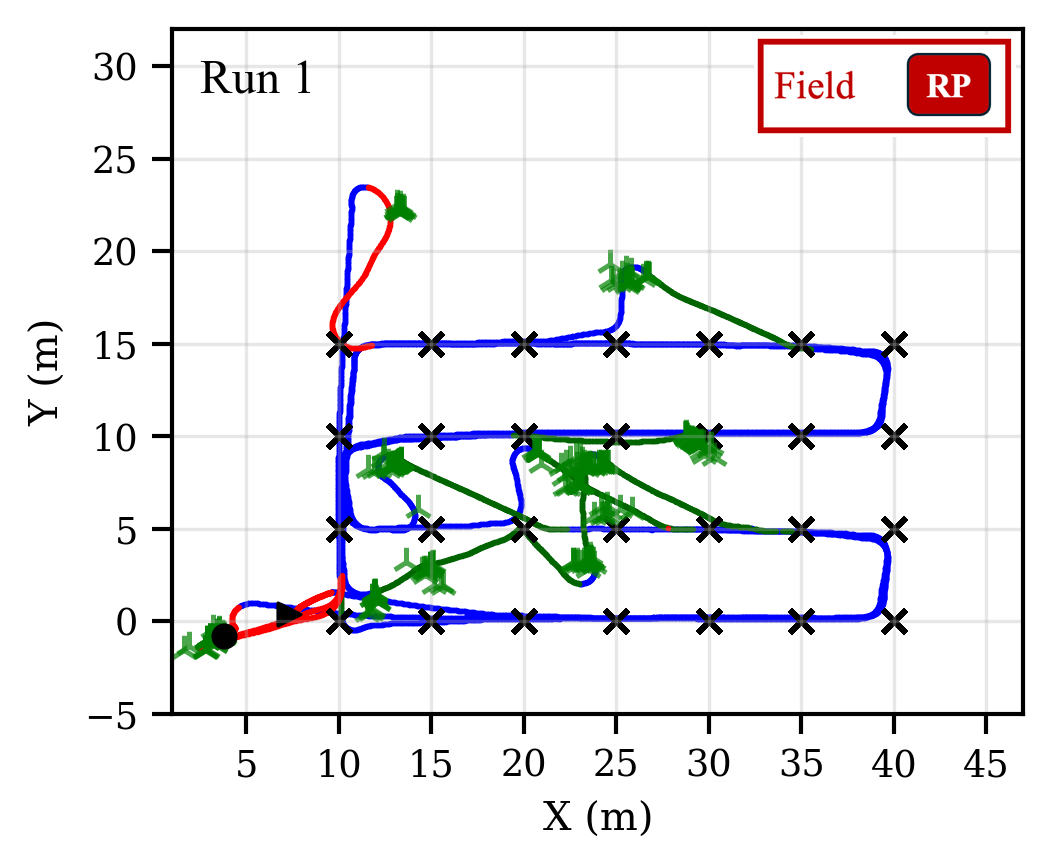}
    \end{subfigure}\hfill
    \begin{subfigure}[b]{0.32\textwidth}\centering
        \includegraphics[width=\linewidth]{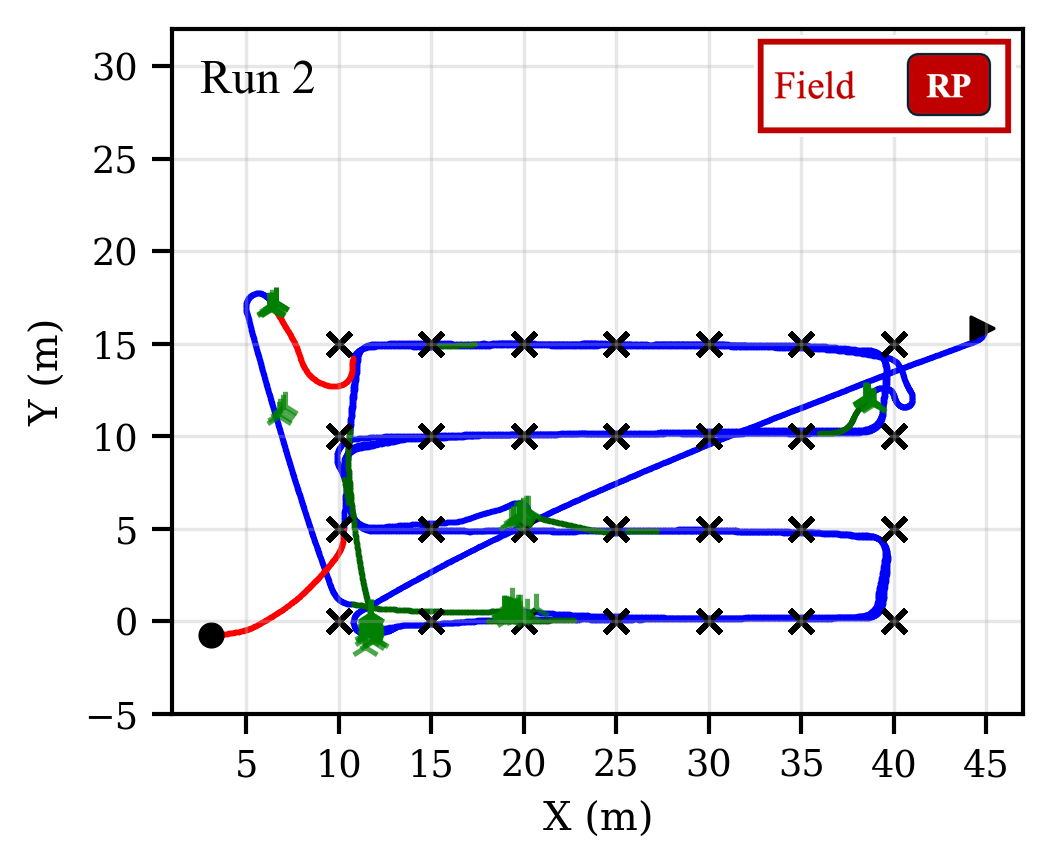}
    \end{subfigure}\hfill
    \begin{subfigure}[b]{0.32\textwidth}\centering
        \includegraphics[width=\linewidth]{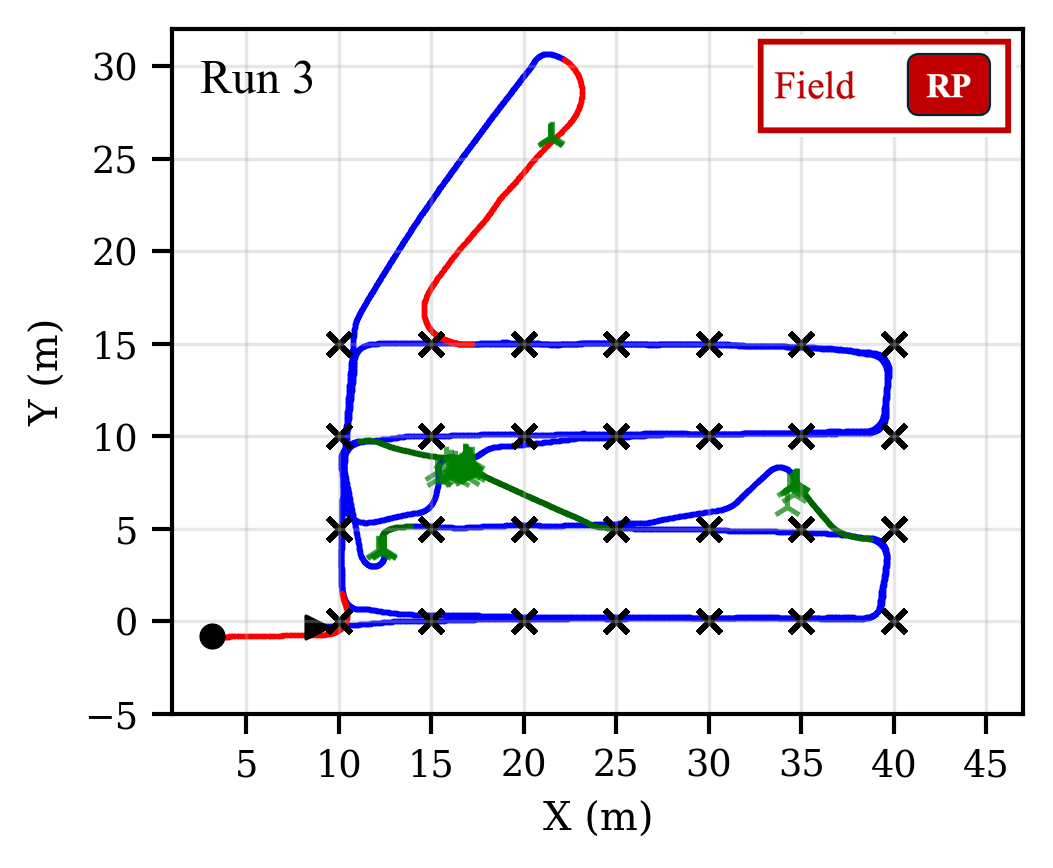}
    \end{subfigure}\hfill

    \begin{subfigure}[b]{1.0\textwidth}\centering
        \includegraphics[width=0.92\linewidth]{figures/waypoints-legend-v.png}
    \end{subfigure}

    \caption{\emph{E3-Real-Perception}: Field trajectories with camera-based perception in the loop. \chg{Each run shows one manual intervention to capture an out-of-bounds target.} All other bottles are collected autonomously.}
    \label{fig:waypoints-real-bottles}
\end{figure*}

We conducted 5 field trials (2 \emph{Small}, 3 \emph{Large}), placing a total of 40 bottles. 
\chg{\Cref{fig:waypoints-real-bottles} shows the three runs in the \emph{Large} grid.
In the first run, 9 bottles were collected autonomously, with a single manual intervention required after one target drifted beyond the predefined area. The manual return to shore for unloading is also visible near the origin. Runs 2 and 3 each show trajectories with 5 captured bottles and required one manual recovery each for a target that moved outside the operational region.
The field-trial statistics are summarized in \cref{tab:integration-metrcis}. Note that for camera-based perception, total distance and completion time are not directly comparable across different runs because bottle locations varied significantly. Small drifts were typically compensated while the bottle remained in the camera field of view, since the target position was updated online. Overall, manual captures were required only when wind-driven drift pushed a bottle beyond the operational area. These results indicate that the perception-to-control loop remains functional under real-world conditions using camera-based perception, but also show that a more robust integration layer, with target memory and drift-aware goal management, is needed to improve capture reliability.}

\subsubsection{Perception Diagnostics}
\begin{figure*}[t]
    \centering
    \includegraphics[width=0.95\textwidth]{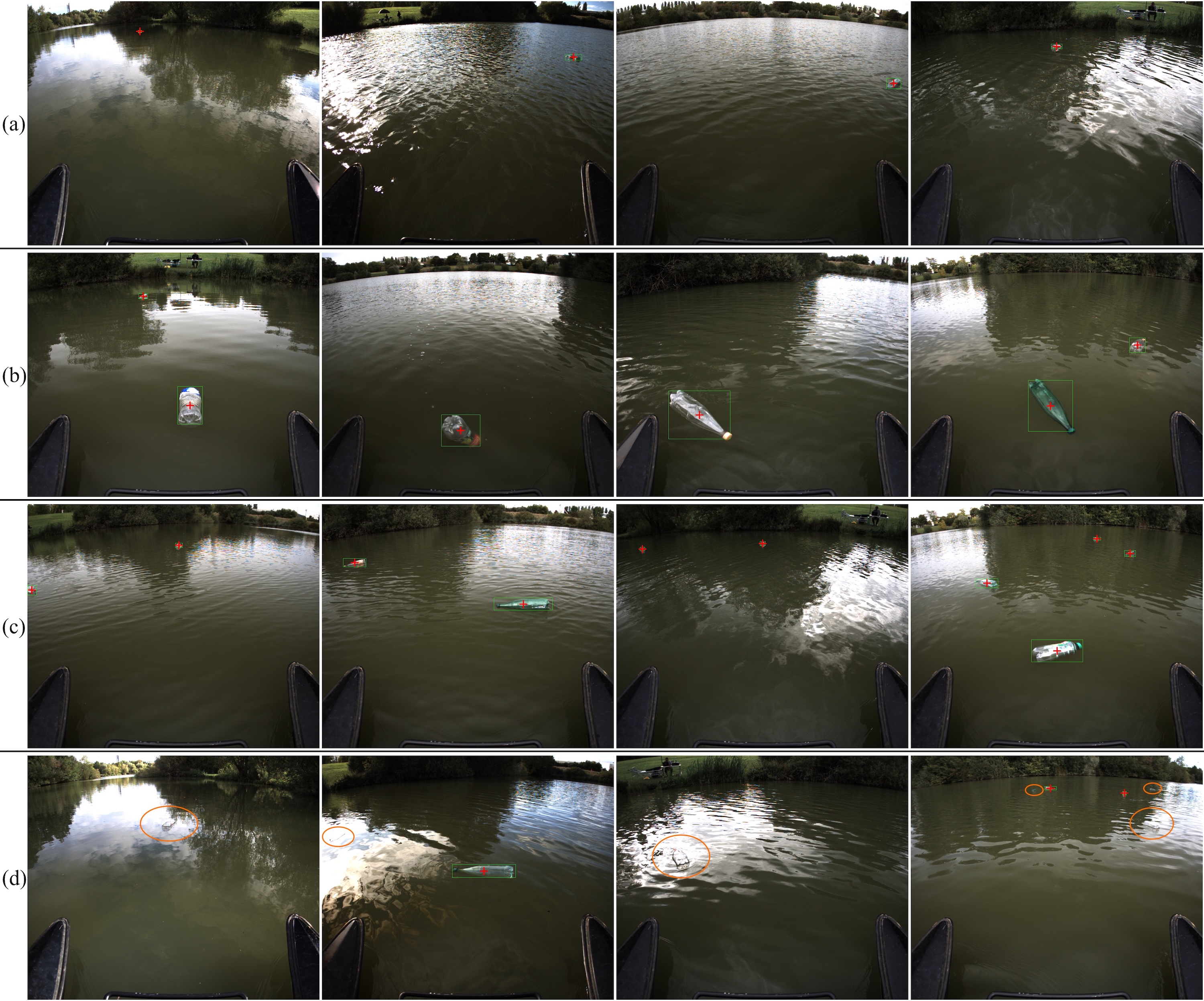}
    \caption{Representative bottle-detection scenarios during autonomous collection. (a) Distant targets detection. (b) Approach phase with targets near the image center before capture. (c) Multi-target scenes, where following one bottle may lose others. (d) Missed detections (orange) caused by saturation or low-contrast, long-range targets.}
    \label{fig:detection-samples}
\end{figure*}

\Cref{fig:detection-samples} illustrates typical visual conditions encountered by our onboard detector during autonomous surface collection. The first row (\textbf{a}) shows early “search” moments in which floating bottles are still distant and occupy only a few pixels. After the vehicle closes in, the second row (\textbf{b}) shows the “approach” phase, with targets steered toward the camera’s principal axis just prior to capture. The third row (\textbf{c}) depicts multi-target scenes, a challenging scenario for the memory-less integration module, where going after one bottle may cause others to exit the field of view. Finally, the fourth row (\textbf{d}) presents missed detections most often associated with long range, low contrast, or sensor saturation due to sun glare.

\begin{figure}[htb]
    \centering
    \includegraphics[width=\linewidth]{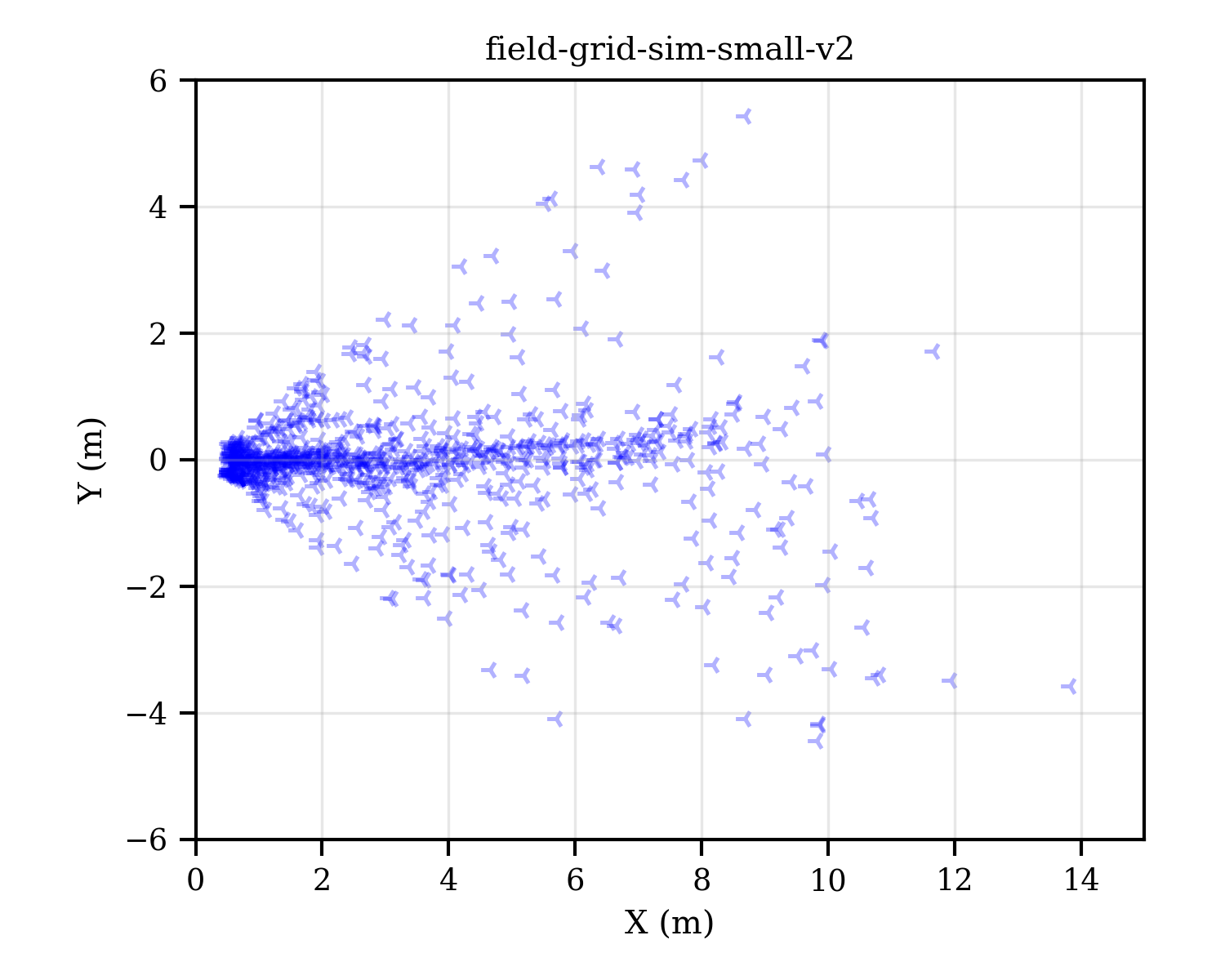}
    \caption{Aggregate distribution of bottle detections in the ASV frame. (Total of 805 detections).
    }
    \label{fig:bottle-heatmap}
\end{figure}

To visualize perception distribution, \cref{fig:bottle-heatmap} aggregates $805$ detections in the ASV coordinate frame. Detections concentrate near the camera’s principal axis with a maximum observed range of $\sim14\,\mathrm{m}$. This spatial distribution is consistent with the expected geometry of the field of view.

\subsubsection{Failure Modes and Limitations}
The observed limitations primarily stem from the simple, memoryless integration policy. Two recurring challenges are identified: (i) concurrent detections and a policy that prioritizes only the most recent and closest detection can lead to goal oscillations when multiple objects are within the camera's field of view; and (ii) targets detected near the field-of-view boundary may drop out before capture. \chg{In field conditions, wind and surface current can also cause the bottle to drift. Small drifts are compensated while the bottle remains in the camera field of view, since the target position is updated online. However, because the integration constrains detours relative to the current waypoint, a drifting bottle may leave the eligible target region before the ASV reaches it, leading to a missed autonomous capture and, in some runs, manual recovery.} The memoryless integration makes recovery inefficient: the ASV may loop through remaining waypoints and restart until a missed target re-enters the field of view.

\subsubsection{Main insights}
Across simulation and field trials, integrating camera-based detections with an RL point-goal controller produced reliable capture behavior, and the perception abstraction enabled identical experiments in simulation and the field, isolating perception integration effects. Together, these experiments validate the control-perception integration. The principal limitations stem from the simplified goal selection policy: favoring the newest or closest detection can induce goal oscillations under concurrent targets and near FoV edges, with impacts increasing under larger search areas, higher target density, and restricted FoV. Small calibration biases (e.g., camera pitch) introduce projection offsets that can perturb goal selection and trajectories, yet their effect on the trained controller was negligible within the tested regimes. These findings suggest that future work should prioritize improvements in the planning layer over changes to the controller itself.

\section{DISCUSSION}
\label{sec:discussion}

\subsection{Limitations and Future Work}

One significant shortcoming of our RL-based control module is the absence of explicit obstacle avoidance within the learned policy. While we already started exploring obstacle avoidance in other preliminary studies, these policies remain untested in field experiments.

One of the main limitations observed during integration was the simplistic strategy for managing multiple targets. While this design served the purpose of evaluating the control under perception uncertainty, it occasionally led the goal selection to prioritize a single target and then 'forgetting' previously observed targets as they disappeared the field of view of the camera. Including improved goal management and long-term tracking must be prioritized in future work.

Such target management could also benefit from reduced projection errors. While these projection errors did not significantly impact the control policy, large range variations complicate the association and tracking of detections. A practical next step is to integrate inertial measurements to compensate pitch variations and to enable online recalibration using the distribution of detections during field tests.

While its impact on our experiments was limited, the perception pipeline would benefit from stronger detection capability. Fine-tuning the model on newly captured data could provide more generalizable detection models with improved performance. In the scope of this evaluation, we trained models restricted to plastic bottles in the dataset. This facilitated validation of the proposed perception and control integration, but is insufficient for real cleanup tasks, where floating debris types are considerably more diverse.


\subsection{Lessons Learned}
\label{subsec:lessons-learned}

This section details practical insights from our deployments. For each topic, we outline key observations, the actions taken, and our approach to addressing specific challenges.

\subsubsection{Sim-to-Real Transferability}
Reliable transfer from simulation to the field is nontrivial because dynamics, sensing latency, and environmental disturbances rarely match their simulated counterparts.
We obtained better transfer by balancing three aspects: (i) increase simulation fidelity where the policy is dominantly sensitive, (ii) apply mild domain randomization on nuisance factors, and (iii) run iterative field trials, isolating components for validation to identify gaps and prioritize based on impact.
The fidelity requirement for zero-shot transfer is task dependent. In our case, improving actuation dynamics modeling contributed more to policy robustness than refining observation noise. This prioritization should not be overgeneralized; however, the workflow of isolating and ranking error sources by observed policy sensitivity can be consistently helpful across different domains.

\subsubsection{Thruster Modeling}
Thruster mismodeling produced policies that were either ineffective or overly conservative. To mitigate this issue, understanding low level actuation signals was critical. In the case of ASVs, separating the effects of thruster dynamics from hull hydrodynamics solely based on field data can be difficult. It makes system identification particularly challenging, especially when having a proprietary microcontroller in the control loop. In our case, the efficient solution was to probe the PPM signal delivered to the motor controller while varying input commands sent to the MCU. This allowed to estimate the embedded control behavior. Incorporating the observed rate-limit into the simulator substantially improved the quality of our actuation model and the performance of the trained policies.

\subsubsection{Simulation Environment}
We found it invaluable to maintain a secondary simulation environment (Gazebo) with independent physics implementation from the learning environment (Isaac Sim). This cross validation of trained policies in a second simulation environment helped expose policy transfer limitations early, allowing fast iteration and supported systems integration checks. It also simplified tasks such as ROS integration and was particularly useful in developing integration module, which reduced the need for time consuming field tests and facilitated solving implementation issues such as frame transforms and threshold tuning.

\subsubsection{Perception Abstraction}
Evaluating a full end-to-end stack complicates error attribution. We introduced a simulated perception abstraction that reuses the calibrated camera model to project water-plane targets. As an intermediary step for modular testing, this enabled fair sim-to-real comparisons and systematic debugging. This was particularly useful to isolate how projection errors influence goal selection before testing with non-reproducible camera detections that lacks ground truth target position.

\subsubsection{Policy Training}
\label{subsubsec:reward-shaping}
Reward design for reinforcement learning is delicate; inappropriate scales or competing terms can stall learning or create unintended behaviors. Goal progress rewards such as approaching and facing the target, were scaled to dominate total reward so the agent learned the task quickly. Secondary shaping terms, such as reducing energy usage, were introduced as penalties and kept an order of magnitude smaller to shape behavior without dominating the combined reward. Without this separation, training often failed to converge or converged to undesired behaviors.

\chg{
\subsubsection{Online Execution of Learned Policies}
\label{subsubsec:onlineexec}
\chg{Once trained, a lightweight RL policy with only two hidden layers of 64 units can run with low latency; in our case, inference is computed on the order of $50\,\mu\mathrm{s}$ on the Jetson Xavier CPU. This shifts most of the computational cost to offline training. This can be an advantage relative to approaches such as MPC, which typically require solving an optimization problem at each control step and can therefore be more computationally demanding during online execution.}
}

\subsubsection{Time Synchronization}
System implementation as multiple independent modules (perception, localization, control) introduce timing errors that can masquerade as controller faults. Latency must be treated carefully with attention to synchronized clocks, message timestamps and frame updated. In our system, perception produced goals at a lower rate and with non-negligible latency. Storing detections in the global frame and re-projecting to the local frame at the control rate using the latest pose reduced temporal inconsistency and removed the need for a dedicated target tracker.

\subsubsection{Perception in Aquatic Environments}
Camera-based perception in outdoor water surface is disrupted by specular reflections, sensor saturation, and rapidly changing illumination, which can create temporarily blind regions in the image plane. We used a polarimetric camera that suppressed reflections where the sensor was not saturated, which can improve detection precision but can't recover from saturated pixels. Since our controller experiments indicated strong robustness against position error in object detection, we consider that sensor with higher dynamic range or improved auto exposure should be prioritized.

\subsubsection{Field Logistics}
Outdoor testing is constrained by weather, battery capacity, limited displays, and access to shore infrastructure, which increases the difficulty of fixing issues during in the field.
To mitigate those issues, we prioritized automating tests, using configuration files and implementing modules that could be reused directly in both simulation and field tests. This approach was especially important for the ablation study with repetitive runs, allowing to validating the full pipeline in simulation before, reducing integration surprises, and increasing productivity during field trials.

\section{CONCLUSION}
\label{sec:conclusion}

This work presented a field-validated perception-to-control architecture \chg{and evaluation methodology} for autonomous surface vessels collecting floating debris. \chg{A careful evaluation protocol, based on dual simulation and perception abstraction, enabled characterization of the sim-to-real gap and identification of the dominant transfer-limiting errors.} \chg{Although developed for an ASV, the underlying methodology may generalize to other mobile robots.} Field experiments under multiple disturbance regimes \chg{show that the controller maintains} high success rates and terminal accuracy under natural wind and current. The main determinant of transfer performance is actuation fidelity, as inaccurate actuator modeling leads to substantial degradation across all metrics. In contrast, substantial \chg{image-plane detection noise} has limited effect on the evaluated control metrics, indicating that, for the point-goal control task considered here, the controller is relatively insensitive to the tested perception errors. \chg{Thus, for this task, the main motivation for polarimetric imaging is glare attenuation, while potential gains in detection position accuracy offer limited downstream benefit for control.} Integrated search-and-capture missions with camera-based detections further confirm end-to-end viability, while the remaining failure modes are mainly associated with memoryless single-target arbitration and limited field of view. Future work should incorporate obstacle avoidance, memory-aware multi-target management, and IMU-assisted online pitch compensation to further improve real-world robustness.

\section*{ACKNOWLEDGMENT}
We would like to thank the members of the DREAM Lab who assisted with field testing and data collection. We also gratefully acknowledge Prof. Seth Hutchinson for his guidance, feedback, and support throughout the development of this work.

\bibliographystyle{IEEEtran}
\bibliography{IEEEabrv,references}

\vfill\pagebreak

\end{document}